\documentclass{article}
\usepackage{iclr2027_conference,times}

\usepackage{amsmath,amsfonts,bm}

\def\eqref#1{equation~\ref{#1}}
\def\1{\bm{1}}

\DeclareMathAlphabet{\mathsfit}{\encodingdefault}{\sfdefault}{m}{sl}
\SetMathAlphabet{\mathsfit}{bold}{\encodingdefault}{\sfdefault}{bx}{n}

\usepackage{amsmath,amssymb,amsfonts,amsthm}

\usepackage{booktabs}
\usepackage{multirow}
\usepackage{graphicx}
\usepackage{subcaption}
\usepackage{array}
\usepackage{colortbl}
\usepackage{wrapfig}

\usepackage{hyperref}
\usepackage{url}
\usepackage{xcolor}
\usepackage{algorithm}
\usepackage{algorithmic}
\usepackage{enumitem}
\usepackage{float}
\usepackage{xspace}

\definecolor{mydarkblue}{rgb}{0,0.08,0.45}
\hypersetup{
    colorlinks=true,
    linkcolor=mydarkblue,
    citecolor=mydarkblue,
    filecolor=mydarkblue,
    urlcolor=mydarkblue
}

\title{From Trajectories to Grounded Preferences: Process Preference Synthesis via Interaction Element Graphs for Web PRMs}

\author{
    Yangzhe Peng\textsuperscript{1}, \quad
    Xiaoyang Wang\textsuperscript{2}, \quad
    Yiyang Zhao\textsuperscript{3,4}, \quad
    Lijun Wu\textsuperscript{2}, \quad
    Kun He\textsuperscript{1} \\
    \textsuperscript{1}Huazhong University of Science and Technology \quad
    \textsuperscript{2}Shanghai Artificial Intelligence Laboratory \quad \\
    \textsuperscript{3}Zhejiang University \quad
    \textsuperscript{4}Shanghai Innovation Institute \\
}

\newcommand{\methodname}{surfPRM}
\newcommand{\method}{\textsc{\methodname}\xspace}
\newcommand{\methoddata}{\textsc{\methodname-data}\xspace}
\newif\ifisarxiv
\isarxivtrue % 开关：设置为 \isarxivtrue 开启 arXiv 预印本模式；设置为 \isarxivfalse 保持匿名评审模式

\ifisarxiv
  \iclrfinalcopy
  \newcommand{\anonymousrepo}{\url{https://github.com/PoloWitty/surfPRM}}
\else
  \newcommand{\anonymousrepo}{\url{https://anonymous.4open.science/r/surfPRM-6188/}}
\fi

\begin{document}

\maketitle
\ifisarxiv
  \lhead{Preprint. Under review.} % 手动覆盖掉原版写死的 Published 字样
\fi

\begin{abstract}
% background
Comparative Process Reward Models (PRMs) provide critical step-level guidance for autonomous web agents by evaluating state-conditioned preferences between candidate actions.
% problem
However, existing preference training data synthesized via multi-policy sampling suffers from a severe scarcity of Grounded Minimal Contrastive Pairs (GMCPs)—where competing candidates target genuine on-page elements with identical action types. In representative baselines preference data (namely, WebArbiter), GMCPs account for merely 24.19\%, biasing PRMs during training to rely on shallow shortcuts (such as element hallucinations and action-type mismatches) rather than acquiring genuine contextual decision semantics.
% method
To address these challenges, we propose \method, a graph-guided process preference synthesis framework for comparative Web PRMs. \method structures web demonstrations into a persistent \textit{Interaction Element Graph} that acts as an environment-grounded negative action proposal mechanism, systematically synthesizing contrastive negative actions across spatial, temporal, and spatiotemporal confusion axes. This elevates the GMCP proportion from 24.19\% to 74.60\%, producing the curated \methoddata dataset.
% exp result
Across six open-source backbones (3B to 9B parameters), PRMs trained on \methoddata outperform baseline-trained models on average on \textsc{WebPRMBench} and rival leading proprietary LLMs. In downstream reward-guided trajectory search on \textsc{WebArena-Lite}, \method provides step-level guidance for both GPT-4o (+14.21\%) and GPT-4o-mini (+12.83\%) policies, yielding substantial improvements in complex web task success rates. Our code and dataset are anonymously available at \anonymousrepo.
\end{abstract}
% TL;DR: While comparative Web PRMs provide critical step-level guidance for autonomous agents, existing preference synthesis via multi-policy sampling suffers from shallow shortcuts like element hallucinations and action-type mismatches problems. We propose surfPRM, a graph-guided framework that structures demonstrations into an Interaction Element Graph to propose environment-grounded, minimally contrastive negative actions. Across six open-source backbones, surfPRM-trained models consistently outperform baselines on WebPRMBench and significantly boost downstream agent search on WebArena-Lite.
% TL;DR (<250 chars): While Web PRMs guide agents, multi-policy preference synthesis introduces shallow shortcuts; we propose surfPRM, synthesizing graph-guided, grounded contrastive pairs to outperform baselines and boost downstream trajectory search performance.

\section{Introduction}
\label{sec:intro}

Autonomous web agents powered by Large Language Models (LLMs) have demonstrated substantial potential in automating complex, multistep workflows across dynamic websites~\citep{deng2023mind2web,zhou2024webarena,he2024webvoyager}. Unlike static text environments, web interaction presents distinct challenges: observation spaces are vast and heterogeneous, action spaces are combinatorially large, and transitions are frequently irreversible (e.g., submitting forms, committing transactions, or navigating stateful applications)~\citep{zhou2024webarena}. In such settings, unguided greedy decoding frequently leads to irreversible dead ends. Step-level verification and test-time search mechanisms are therefore valuable to guide agents toward successful task completion~\citep{koh2025treesearch,yang2026gta1}.

Conventional Outcome Reward Models (ORMs)~\citep{ouyang2022training,rafailov2023direct} evaluate agent trajectories solely upon completion, providing delayed and sparse scalar rewards. This outcome-centric evaluation creates acute credit assignment ambiguity: an ORM cannot identify which intermediate step derailed execution, nor can it detect when an agent enters a dead end mid-trajectory. Process Reward Models (PRMs)~\citep{uesato2022solving,lightman2024lets,wang2024mathshepherd} address this by verifying steps incrementally. In the web domain, state-of-the-art Web PRMs adopt a \textit{comparative, generative} formulation~\citep{zhang2026webarbiter}. Given state $s_t = (I, O_t, H_t, u_t)$—instruction $I$, linearized Accessibility Tree (AXTree) observation $O_t$, action history $H_t$, and URL $u_t$—a comparative Web PRM evaluates candidate actions $(a_t^1, a_t^2)$ conditioned on $s_t$: $f_\theta(s_t, a_t^1, a_t^2) \to (j_t, \hat{y}_t)$ with $\hat{y}_t \in \{1, 2\}$, where $j_t$ is an explicit \textit{judgment rationale} detailing causal distinctions between candidates, and $\hat{y}_t$ indicates the preferred candidate index ($\hat{y}_t = 1 \iff a_t^1 \succ a_t^2$). Rather than predicting uncalibrated absolute scores across disparate web layouts, the comparative PRM learns relative preferences between competing actions under the exact same state~\citep{chae2025webshepherd,zhang2026webarbiter}.

% 监督形式 → 现有采样缺陷 → 局部过滤规则局限性
To train such comparative verifiers, supervision takes the form of process preference tuples $d_t = (s_t, a_t^+, a_t^-, j_t^*, y_t^*)$. In preference synthesis, successful demonstration trajectories naturally provide execution steps that serve as \textit{reference positives} ($a_t^+$). Given demonstration steps as reference positives, constructing environment-grounded and informative alternatives ($a_t^-$) becomes a central challenge in comparative Web PRM data synthesis. Previous pipelines~\citep{zhang2026webarbiter,chae2025webshepherd} typically sample alternative actions $a_t^-$ via multi-policy models and prompt teacher LLMs to synthesize comparative rationales. However, existing preference training data synthesized via multi-policy sampling suffers from a severe scarcity of Grounded Minimal Contrastive Pairs (GMCPs)—where competing candidates target genuine on-page elements with identical action types. In representative baseline data (namely, WebArbiter~\citep{zhang2026webarbiter}), GMCPs account for merely 24.19\%, with the remaining majority dominated by shallow shortcuts: (1)~\textit{Element Hallucination}, targeting element identifiers (BIDs) absent from $O_t$; and (2)~\textit{Action-Type Mismatch}, diverging in action modality (e.g., valid \texttt{click} vs. invalid \texttt{fill}). This severe deficit of minimal contrastive pairs biases PRMs during training to rely on superficial cues rather than acquiring genuine contextual decision semantics. Naively filtering multi-policy sampled actions against active AXTree elements fails to resolve this dilemma: it discards the vast majority of generated candidates—incurring substantial LLM sampling costs with low yield of usable pairs—while failing to exploit the rich interaction transitions already embedded in demonstration trajectories.

To address these challenges, we propose \method, a \textit{Graph-guided Process Preference Synthesis} framework. We aggregate web interaction demonstrations into a persistent \textit{Interaction Element Graph} $G = (V, E_s, E_\tau)$, where nodes represent persistent interactive elements, $E_s$ captures AXTree structural containment within pages, and $E_\tau$ captures temporal interaction transitions across successful tasks. Crucially, the graph functions as an environment-grounded \textit{negative action proposal mechanism} $a_t^- \sim q_G(a \mid s_t, a_t^+)$, proposing contrastive candidates across spatial (structural), temporal, and spatiotemporal confusion axes. A strong teacher model then filters false negatives via multi-trial consistency and synthesizes authoritative judgment rationales, yielding the high-quality \methoddata preference dataset. Our core contributions are: (1)~\textbf{Problem Diagnosis}: We identify the severe scarcity of Grounded Minimal Contrastive Pairs (GMCPs) as a key deficiency in existing multi-policy sampling Web PRM data synthesis pipelines, which introduces shallow shortcuts that bias PRMs during training to rely on superficial cues rather than contextual decision semantics. (2)~\textbf{Graph Proposal Mechanism}: We introduce the Interaction Element Graph proposal mechanism in \method with three-axis confusion rules, elevating the GMCP proportion from 24.19\% in WebArbiter to 74.60\% in \methoddata while preserving legitimate non-DOM actions. (3)~\textbf{Empirical Validation}: Across six open-source backbones (3B to 9B parameters), comparative Web PRMs trained on \methoddata outperform baseline-trained models on average on \textsc{WebPRMBench} and rival leading proprietary LLMs, delivering +14.21\% (GPT-4o) and +12.83\% (GPT-4o-mini) success gains in downstream \textsc{WebArena-Lite} search.

\section{Related Work}
\label{sec:related_work}

\paragraph{Web Agents and Interactive Decision-Making.}
LLM-powered web and computer-use agents have advanced across benchmarks such as WebShop~\citep{yao2022webshop}, Mind2Web~\citep{deng2023mind2web}, WebArena~\citep{zhou2024webarena}, WebVoyager~\citep{he2024webvoyager}, VisualWebArena~\citep{koh2024visualwebarena}, and OSWorld~\citep{xie2024osworld}. To support learning in these environments, prior work has scaled demonstration synthesis~\citep{ou2024synatra,xu2025agenttrek,murty2025nnetnav,pahuja2025explorer,sun2025osgenesis,wang2026synthagent} and developed synthetic environments and environment models~\citep{gao2026websynthesis,fang2025webevolver,xiao2026webworld,zhang2026infiniteweb,fan2026webfactory,wu2026autowebworld,chae2026verienv,bai2026webgym}. However, web interfaces present dense, visually and semantically overlapping interactable elements, where empirical failure analyses identify low-level execution and grounding errors—rather than high-level planning—as the primary bottleneck~\citep{aghzal2026why}. Because imitation learning on positive demonstrations only exposes agents to successful paths, it provides no discriminative signal against superficially plausible decoy elements in the same observation, highlighting the need for fine-grained step-level verification and evaluative feedback to distinguish competing actions.

\paragraph{Web Agent Reward Models.}
Reward modeling for web agents initially focused on trajectory-level Outcome Reward Models (ORMs), such as the outcome evaluators in WebRL~\citep{qi2025webrl} and WebAgent-R1~\citep{wei2025webagentr1}, which assess overall task completion to guide reinforcement learning. However, trajectory-level feedback provides limited guidance for attributing outcomes to individual intermediate actions~\citep{lu2025agentrewardbench}. Web Process Reward Models (PRMs) instead provide step-level guidance: WebShepherd~\citep{chae2025webshepherd} constructs a benchmark with 40K preference pairs using multi-policy sampling and heuristic filtering, while WebArbiter~\citep{zhang2026webarbiter} generates preference verdicts with principle-guided judgment rationales. These methods use policy-generated candidates, which do not by themselves ensure grounded, action-type-controlled comparisons; including candidate rationales can additionally expose stylistic cues. Our analysis of WebArbiter data finds that Grounded Minimal Contrastive Pairs (GMCPs) account for only 24.19\% of pairs. We focus on constructing process preference supervision: in \method, an Interaction Element Graph proposes contrastive negative actions for successful demonstration steps, followed by teacher filtering and synthesis of judgment rationales and preference verdicts for comparative Web PRM training.

\paragraph{Process Supervision and Step-Level Data Synthesis.}
Originating in mathematical reasoning, process supervision frameworks such as Math-Shepherd~\citep{wang2024mathshepherd} automated step-level annotation via Monte Carlo (MC) rollouts from intermediate states, while ProcessBench~\citep{zheng2025processbench} showed that PRM generalization depends crucially on data quality rather than raw volume. However, translating rollout-based process supervision to web environments is fundamentally intractable: live web pages cannot be arbitrarily cloned or reset, actions such as form submission are irreversible, and automated outcome verification is brittle~\citep{aghzal2026why,lu2025agentrewardbench}. Although both domains address step-level credit assignment, recent agentic RL methods (e.g., GiGPO~\citep{feng2025gigpo}, GraphGPO~\citep{cheng2026graphgpo}) operate under an orthogonal paradigm focused on online policy optimization through interactive exploration. In contrast, \method constructs an comparative verifier and process reward model to serve as an evaluative guide for downstream inference-time search. To synthesize reliable process preference supervision without requiring active environment rollouts or resets, \method perturbs golden demonstration steps along spatial, temporal, and spatiotemporal axes on a persistent Interaction Element Graph, constructing environment-grounded, minimally contrastive negatives from existing trajectory graphs.

\section{Method}
\label{sec:method}

\begin{figure}[t]
\centering
\includegraphics[width=0.96\textwidth,height=3.6cm,keepaspectratio]{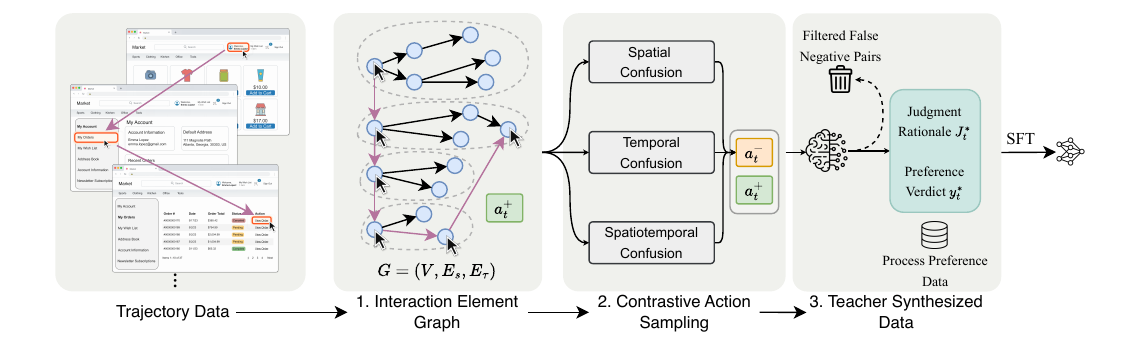}
\caption{\textbf{Overview of the \method Framework.} Demonstration trajectories are aggregated into an Interaction Element Graph $G=(V, E_s, E_\tau)$ capturing spatial containment and temporal paths. The graph acts as a Negative Action Proposal Mechanism across spatial, temporal, and spatiotemporal axes. Teacher models filter false negatives and synthesize gold rationales and verdicts for comparative PRM training.}
\label{fig:pipeline_overview}
\vspace{-1.5em}
\end{figure}

\subsection{Problem Formulation and Comparative PRM Pipeline}
\label{sec:method_formulation}

\paragraph{State, Action, and Comparative PRM Formulation.}
We formulate web navigation as a sequential decision process over discrete steps $t \in \{1, \dots, T\}$. At step $t$, the environmental state is $s_t = (I, O_t, H_t, u_t)$, where $I$ is the user instruction, $O_t$ is the linearized Accessibility Tree (AXTree) with discrete element identifiers ($\text{BID} \in \mathbb{N}$), $H_t = (a_1, \dots, a_{t-1})$ is the action history, and $u_t$ is the current URL. An action $a_t = (\mathrm{op}_t, \text{target}, \text{value})$ specifies the operation type $\mathrm{op}_t \in \{\texttt{click}, \texttt{fill}, \texttt{hover}, \texttt{select\_option}, \dots\}$, target element identifier, and optional parameters. A comparative Web PRM $f_\theta$ evaluates two candidate actions $(a_t^1, a_t^2)$ conditioned on $s_t$: $f_\theta(s_t, a_t^1, a_t^2) \to (j_t, \hat{y}_t)$ with $\hat{y}_t \in \{1, 2\}$, where $j_t$ is an explicit \textit{judgment rationale} providing comparative causal reasoning, and $\hat{y}_t$ indicates the preferred candidate index ($\hat{y}_t = 1 \iff a_t^1 \succ a_t^2$)~\citep{zhang2026webarbiter}. Under this setting, PRM training data consists of preference tuples $d_t = (s_t, a_t^+, a_t^-, j_t^*, y_t^*)$, where $a_t^+$ is the positive step from an expert demonstration, $a_t^-$ is a contrastive negative action, $j_t^*$ is the gold judgment rationale, and $y_t^*$ is the ground-truth verdict.

\begin{figure}[t]
\centering
\includegraphics[width=0.92\textwidth]{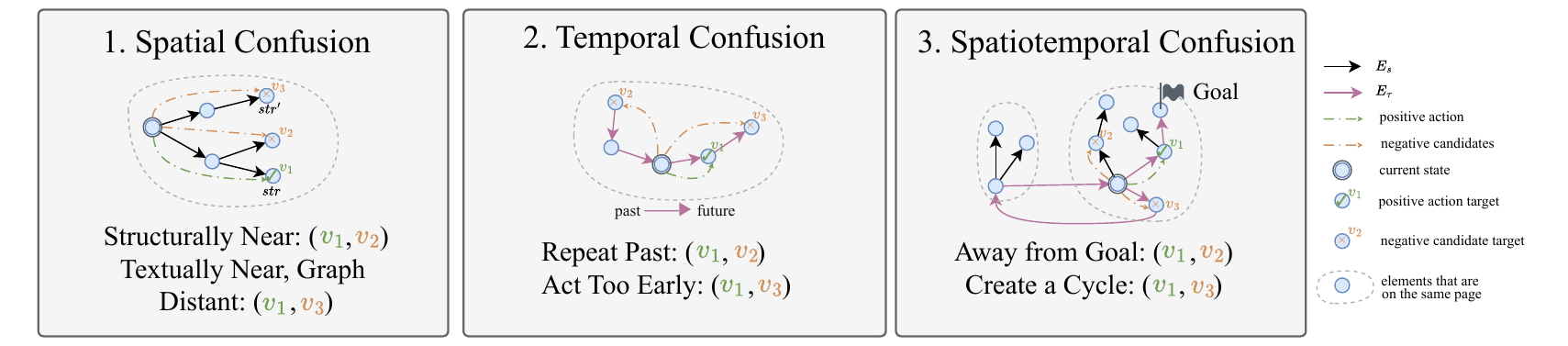}
\caption{\textbf{Negative Action Synthesis across Three Confusion Axes.} Candidate negatives are proposed from the Interaction Element Graph along spatial, temporal, and spatiotemporal axes, targeting distinct confusion modes (formal definitions in Appendix~\ref{sec:app_rules}).}
\label{fig:neg_gen}
\vspace{-1.5em}
\end{figure}

\subsection{Website Interaction Element Graph Construction}
\label{sec:method_graph}

To provide structured proposals for negative actions, \method aggregates demonstration trajectories into persistent, site-level \textbf{Interaction Element Graphs} $G = (V, E_s, E_\tau)$.

\paragraph{Persistent Element Identity Mapping.}
Runtime Backend Node IDs (BIDs) in AXTrees are ephemeral across reloads and sessions. To track elements across trajectories, we map each element to a persistent node $v = \phi(\text{site},\, \text{URL},\, \text{elem\_ctx}) \in V_{\text{elem}}$ using BID-invariant context (e.g., accessible role and name, ancestor chain, sibling position, and subtree signature; details in Appendix~\ref{sec:app_action_mapping}). For actions without DOM anchors (\texttt{scroll}, \texttt{go\_back}, \texttt{new\_tab}, \texttt{respond}), we instantiate page-scoped Virtual Nodes $v_{\text{virt}} \in V_{\text{virt}}$ keyed by the landing page and action type. Instantiating virtual nodes per page rather than globally preserves local topological fidelity, yielding the unified node set $V = V_{\text{elem}} \cup V_{\text{virt}}$.

\paragraph{Graph Topology.}
For each website, the graph $G = (V, E_s, E_\tau)$ captures two complementary topologies:
(1)~\textbf{Spatial Containment Edges ($E_s$)}: On each page observation, nodes represent BID-bearing interactive elements in the AXTree. For each element $v$, we traverse upward along its ancestor chain, skipping inert structural containers (\texttt{listitem}, \texttt{generic}, etc.), and connect an edge $(u, v) \in E_s$ to its \textit{nearest interactive ancestor} $u$, recording the hierarchy gap ($\text{depth\_gap}$), with duplicates removed across observations. Bypassing non-operable containers avoids injecting hundreds of action-free, noisy edges per page. Instead, linking solely to the nearest interactive ancestor directly captures \textit{nested interactive structures} (e.g., options within comboboxes)---inducing realistic scoping ambiguities (inner vs.\ outer targets) critical for contrastive negatives while eliminating transitive redundancy.
(2)~\textbf{Temporal Interaction Edges ($E_\tau$)}: Directed transitions $(v_t \xrightarrow{a_t} v_{t+1})$ observed between consecutive steps across successful trajectories.
The graph reflects observed structural and temporal relationships rather than a complete world model, proposing plausible alternative candidates rather than proving unobserved paths impossible.

\subsection{Graph-Based Grounded Minimal Negative Action Synthesis}
\label{sec:method_neg_synthesis}

Given a positive step $(s_t, a_t^+)$ targeting element $v_t^+$, the \method proposal mechanism samples candidate actions $a_t^- \sim q_G(a \mid s_t, a_t^+)$ from $G$. Importantly, virtual nodes $V_{\text{virt}}$ serve to model trajectory flow and represent non-DOM positive steps (e.g., \texttt{scroll} or terminal responses) paired with competing page-element negatives, but are never proposed as negative actions. To prevent element hallucinations, negative candidates are drawn strictly from $\mathcal{V}_{\text{grounded}}(s_t)$, the set of concrete DOM elements in $G$ uniquely resolvable in the active observation $O_t$; proposed candidate nodes that cannot be uniquely bound to an active BID in $O_t$ are discarded. This mechanism elevates the proportion of Grounded Minimal Contrastive Pairs (GMCPs) to 74.60\%, balancing grounded minimal contrast with the need to preserve legitimate virtual positive steps and respect element affordances, rather than enforcing artificial 100\% GMCP pairing.

We synthesize contrastive negatives across three confusion axes comprising six rules $\mathcal{R} = \{r_1, \dots, r_6\}$ (Figure~\ref{fig:neg_gen}; formal definitions are detailed in Appendix~\ref{sec:app_rules}):
\begin{itemize}[leftmargin=*,itemsep=0pt,topsep=1pt,parsep=0pt,partopsep=0pt]
    \item \textbf{Spatial Confusion}: Perceptual distraction within the active page: (1) \textit{Structurally Near}, selecting a nearby element within a few AXTree containment hops of the positive target (e.g., a nearby date button \textit{``Feb 25''} instead of \textit{``Jan 28''} in a JetBlue booking calendar); and (2) \textit{Textually Near, Graph-Distant}, selecting an element with high text similarity ($>0.75$, character-level sequence-matching ratio) but $>4$ hops away or unreachable in the combined element graph (containment plus interaction edges) (e.g., selecting the 1984 film link instead of the 2021 release for \textit{Dune} on Rotten Tomatoes). Note that graph distance is essential: elements reachable within $\le 4$ hops in the combined graph across trajectories are excluded to avoid penalizing viable multi-step paths.
    \item \textbf{Temporal Confusion}: Sequential misordering along the interaction timeline: (3) \textit{Repeat Past}, repeating an operation from $k$ steps prior ($1 \le k \le N$) still visible in $O_t$ (e.g., repeating a \texttt{hover} action on an artist profile on SoundCloud); and (4) \textit{Act Too Early}, executing an action scheduled $k$ steps ahead ($1 \le k \le N$) that is visible prematurely (e.g., clicking search submit on Amazon before entering keywords).
    \item \textbf{Spatiotemporal Confusion}: Workflow divergence from global task progression: (5) \textit{Away from Goal}, increasing shortest graph distance to the task's final interaction element in the combined element graph (e.g., clicking an unrelated promotional item rather than submitting search on Micro Center); and (6) \textit{Create a Cycle}, transitioning back to a previously visited state and creating a loop in $E_\tau$ (e.g., clicking the home logo on IMDb while viewing search auto-suggestions).
\end{itemize}
The complete proposal distribution combines all active rules: $\mathcal{N}(s_t, a_t^+) = \bigcup_{r \in \mathcal{R}} \mathcal{N}_r(s_t, a_t^+; G)$. To prevent position bias, each pair $(a_t^+, a_t^-)$ is randomly shuffled into candidate positions $(a_t^1, a_t^2)$, with the ground-truth preference verdict $y_t^* \in \{1, 2\}$ dynamically set to track the index of the positive action, represented without candidate rationales~\citep{zheng2023judging,wang2024large} (Appendix~\ref{sec:app_action_only}).

\subsection{Teacher Verification and Gold Supervision Synthesis}
\label{sec:method_teacher_synthesis}

To ensure label reliability and filter potential false negatives (where the negative action represents an equally valid alternative path), we evaluate each candidate pair using a high-capacity Teacher Model over $K=5$ independent trials: $(j_t^{(k)}, \hat{y}_t^{(k)}) \sim \text{Teacher}(s_t, a_t^1, a_t^2)$. A candidate pair is flagged as a false negative and discarded if and only if the teacher contradicts the trajectory label in \textit{every single trial} ($\forall k, \hat{y}_t^{(k)} \neq y_t^*$). Out of 10,000 sampled candidate pairs, only 179 were filtered out under this criterion, yielding a high verification pass rate of 98.21\%.

For each retained pair, we extract a consistent judgment rationale corresponding to the correct label as the \textbf{Gold Judgment Rationale $j_t^*$}. The rationale articulates user intent, active page affordances, and the causal justification for why $a_t^{y_t^*}$ advances the goal while the other action represents a failure mode. Using the curated \methoddata dataset $\mathcal{D} = \{ (s_t, a_t^1, a_t^2, j_t^*, y_t^*) \}_{i=1}^N$, we train the comparative Web PRM $f_\theta$ via auto-regressive cross-entropy loss:
\begin{equation}
    \mathcal{L}_{\text{SFT}}(\theta) = - \sum_{i=1}^N \sum_{k=1}^{|Y_t|} \log P_\theta(y_{t, k} \mid s_t, a_t^1, a_t^2, y_{t, <k})
\end{equation}
where target sequence $Y_t = (j_t^* \circ y_t^*)$.

\section{Data Collection and Dataset Statistics}
\label{sec:data_statistics}

\subsection{Source Trajectories and Graph Construction}
Our dataset, \methoddata, is constructed from successful web interaction trajectories drawn from the WPRM demonstration pool~\citep{chae2025webshepherd}. The source pool spans 50 diverse websites with 1,409 normalized URL paths across e-commerce, content management systems, travel booking, technical forums, and administrative dashboards. We process 854 successful trajectories comprising 9,497 page observations to construct 50 site-level subgraphs. The aggregated graphs encompass 116,182 element nodes ($V$), 1,532 deduplicated Spatial Containment Edges ($E_s$), and 6,739 deduplicated Temporal Interaction Edges ($E_\tau$).
%  derived from 7,958 observed step transitions. 
On average, each site subgraph contains 2,324 element nodes, 134.8 Temporal Interaction Edges, 30.6 Spatial Containment Edges, and 17.1 trajectories. Notably, our source demonstrations match those of WebArbiter~\citep{zhang2026webarbiter}, ensuring performance differences stem solely from the negative action synthesis process.

\vspace{-0.5em}
\subsection{Dataset Curation and Final Statistics}
\label{sec:dataset_curation_stats}

\begin{wraptable}{r}{0.48\textwidth}
\vspace{-1.2em}
\caption{Distribution of Proposal Rules in \methoddata (10k Pairs).}
\label{tab:data_distributions}
\centering
\footnotesize
\setlength{\tabcolsep}{2pt}
\begin{tabular}{llr}
\toprule
\textbf{Proposal Rule} & \textbf{Axis} & \textbf{Count} \\
\midrule
Textually Near, Graph-Distant & Spatial & 5,022 \\
Away from Goal & Spatiotemp. & 4,020 \\
Act Too Early & Temporal & 304 \\
Structurally Near & Spatial & 223 \\
Repeat Past & Temporal & 166 \\
Create a Cycle & Spatiotemp. & 86 \\
\midrule
Total & & 9,821 \\
\bottomrule
\end{tabular}
\vspace{-1.2em}
\end{wraptable}

Graph traversal across the 50 subgraphs produced 31,648 mined candidates. Applying environmental grounding (requiring elements to uniquely resolve in active observation $O_t$) yielded a pool of 31,400 candidates. We sampled 10,000 candidate pairs (seed 42) for teacher verification, matching the data scale of the WebArbiter SFT baseline to enable a controlled comparison at an aligned budget. Following multi-trial teacher consistency filtering ($K=5$), \textbf{9,821 pairs passed verification} (179 pairs filtered, yielding an agreement pass rate of 98.21\%; detailed teacher verification costs are reported in Appendix~\ref{sec:app_teacher_cost}), forming the final \methoddata dataset. Table~\ref{tab:data_distributions} details the distribution of proposal rules. 
% Among them, \textit{Textually Near, Graph-Distant} (5,022) and \textit{Away from Goal} (4,020) account for the majority of proposals. 
The final dataset covers 1,084 normalized URL paths (after host lowercasing, \texttt{www}-prefix stripping, and trailing-slash removal). 

\method increases the proportion of Grounded Minimal Contrastive Pairs (GMCPs) from \textbf{24.19\%} (WebArbiter baseline) to \textbf{74.60\%} (7,326 out of 9,821 pairs), more than a threefold improvement. Crucially, among the remaining 25.40\% non-GMCP pairs, 0\% fail due to target elements being absent from the current page; every proposed negative action is strictly grounded in an active element within $O_t$. Instead, the non-GMCP residual stems from two deliberate design choices: (1)~virtual positive pairs (16.6\%), where demonstration steps whose positive action is a global viewport or navigation operation (e.g., \texttt{scroll} or terminal responses) are paired with competing page-element negatives, diverging in target modality; and (2)~affordance-driven action-type divergence (9.1\%), where negative candidates adopt the operation type historically demonstrated on that element to respect executable affordances (e.g., clicking a link vs.\ filling an input) rather than coercing an identical operation type onto incompatible elements. A full structural decomposition is provided in Appendix~\ref{sec:app_action_mapping}.

\vspace{-0.5em}
\subsection{Quality Assurance and Iterative Refinement}
\label{sec:data_qa}
We instituted automated and rule-based verification protocols to guarantee execution compliance: (1)~\textbf{Contextual Value Completion}: Missing arguments for input-dependent actions (e.g., \texttt{fill}) are contextually completed by a model conditioned on AXTree context, and non-editable controls (e.g., custom comboboxes) are normalized to valid affordances like \texttt{click} (Appendix~\ref{sec:app_qa_details}); audits confirmed zero empty values.
(2)~\textbf{Human False-Negative Audit}: In a manual audit on 100 randomly sampled pairs from both WebArbiter and \methoddata, 30 of 100 WebArbiter negatives were false negatives (valid alternative actions mislabeled as inferior), whereas \methoddata yielded only 12---a $2.5\times$ reduction in false-negative rate ($30\% \to 12\%$), confirming substantially cleaner and more dependable process preference supervision.

\section{Experiments}
\label{sec:experiments}

\subsection{Experimental Setup}
\label{sec:exp_setup}

\begin{table}[t]
\centering
\caption{\textbf{Step-Level Process Preference Discrimination on WebPRMBench.} Evaluated on Pairwise and BoN Accuracy (\%). Average reports macro-average across the three environments. $*$ denotes baseline results directly reported from the \textsc{WebArbiter} paper. Models fine-tuned on \methoddata are shaded; \textbf{bold} marks best within each group in Average.}
\label{tab:web_prm_bench}
\footnotesize
\setlength{\tabcolsep}{3.9pt}
\renewcommand{\arraystretch}{1.0}
\begin{tabular}{lcccccccc}
\toprule
\multirow{2}{*}{\textbf{Model}} & \multicolumn{2}{c}{\textbf{WebArena}} & \multicolumn{2}{c}{\textbf{AssistantBench}} & \multicolumn{2}{c}{\textbf{WorkArena}} & \multicolumn{2}{c}{\textbf{Average}} \\
\cmidrule(lr){2-3} \cmidrule(lr){4-5} \cmidrule(lr){6-7} \cmidrule(lr){8-9}
& Pairwise & BoN & Pairwise & BoN & Pairwise & BoN & Pairwise & BoN \\
\midrule
\multicolumn{9}{l}{\textit{\textbf{Closed-Source LLM}}} \\
% GPT-4o-mini* & 78.23 & 56.72 & 89.17 & 73.33 & 81.43 & 46.70 & 82.94 & 58.92 \\
% GPT-4o* & 84.58 & 66.67 & 85.83 & 66.67 & 84.33 & 55.19 & 84.91 & 62.84 \\
GPT-5* & 84.83 & 71.64 & 81.67 & 63.33 & 81.14 & 64.62 & 82.55 & 66.53 \\
% Claude-3.7-Sonnet* & 82.80 & 64.10 & 81.50 & 61.30 & 82.10 & 60.60 & 82.13 & 62.00 \\
Claude-4.6-Sonnet & 84.58 & 71.14 & 80.00 & 60.00 & 83.02 & 66.51 & 82.53 & 65.88 \\
% Gemini-2.5-Flash* & 82.71 & 62.19 & 80.00 & 63.33 & 83.30 & 56.13 & 82.00 & 60.55 \\
Gemini-3.7-Flash & 83.71 & 68.66 & 83.33 & 73.33 & 78.77 & 64.15 & 81.94 & 68.71 \\
Jev-1.13 (choice mode) & 84.83 & 70.65 & 79.17 & 60.00 & 78.42 & 58.49 & 80.81 & 63.05 \\
Jev-1.13 (noul mode) & 84.70 & 69.65 & 76.67 & 56.67 & 78.89 & 57.55 & 80.09 & 61.29 \\
\midrule
\multicolumn{9}{l}{\textit{\textbf{Open-Source LLM}}} \\
Llama-3-70B-Instruct* & 77.36 & 50.75 & 85.83 & 70.00 & 79.08 & 40.09 & 80.76 & 53.61 \\
DeepSeek-R1* & 82.04 & 60.21 & 78.49 & 56.18 & 84.12 & 63.89 & 81.55 & 60.09 \\
DeepSeek-V4-Flash-0731 & 83.21 & 71.64 & 80.00 & 63.33 & 79.36 & 58.49 & 80.86 & 64.49 \\
\midrule
\multicolumn{9}{l}{\textit{\textbf{Fine-Tuned LLM}}} \\
WebShepherd-3B* & 68.16 & 41.29 & 66.67 & 46.67 & 50.00 & 21.23 & 61.61 & 36.40 \\
WebShepherd-8B* & 68.33 & 43.88 & 55.92 & 30.00 & 54.56 & 25.53 & 59.60 & 33.14 \\
\addlinespace[2pt]
\textbf{Qwen2.5-3B-Instruct}  & 70.02 & 35.82 & 78.33 & 46.67 & 68.99 & 29.25 & 72.45 & 37.25 \\
\quad + WebArbiter 10k SFT & 81.84 & 55.22 & 69.17 & 40.00 & 71.58 & 35.38 & 74.20 & 43.53 \\
\rowcolor{gray!20}\quad + Ours 10k SFT & 81.59 & 55.22 & 85.00 & 56.67 & 75.35 & 41.98 & \textbf{80.65} & \textbf{51.29} \\
\addlinespace[2pt]
\textbf{Qwen2.5-7B-Instruct}  & 73.88 & 44.78 & 81.67 & 50.00 & 72.76 & 37.26 & 76.10 & 44.01 \\
\quad + WebArbiter 10k SFT & 80.35 & 54.23 & 83.33 & 60.00 & 78.07 & 51.42 & 80.58 & 55.22 \\
\rowcolor{gray!20}\quad + Ours 10k SFT & 84.33 & 62.19 & 85.83 & 56.67 & 77.95 & 50.00 & \textbf{82.70} & \textbf{56.29} \\
\addlinespace[2pt]
\textbf{Qwen3-4B}  & 73.38 & 47.26 & 78.33 & 43.33 & 69.93 & 36.79 & 73.88 & 42.46 \\
\quad + WebArbiter 10k SFT & 84.95 & 63.18 & 76.67 & 46.67 & 75.12 & 48.11 & 78.91 & 52.65 \\
\rowcolor{gray!20}\quad + Ours 10k SFT & 83.71 & 60.20 & 80.83 & 56.67 & 79.36 & 51.42 & \textbf{81.30} & \textbf{56.10} \\
\addlinespace[2pt]
\textbf{Qwen3-8B}  & 76.24 & 50.75 & 80.83 & 56.67 & 73.00 & 38.21 & 76.69 & 48.54 \\
\quad + WebArbiter 10k SFT & 84.58 & 64.18 & 80.83 & 60.00 & 75.94 & 48.11 & 80.45 & 57.43 \\
\rowcolor{gray!20}\quad + Ours 10k SFT & 84.58 & 62.19 & 88.33 & 73.33 & 79.72 & 52.36 & \textbf{84.21} & \textbf{62.63} \\
\addlinespace[2pt]
\textbf{Qwen3.5-4B} & 81.72 & 62.69 & 75.83 & 50.00 & 80.07 & 53.77 & 79.21 & 55.49 \\
\quad + WebArbiter 10k SFT & 83.83 & 59.70 & 86.67 & 60.00 & 76.89 & 47.64 & 82.46 & 55.78 \\
\rowcolor{gray!20}\quad + Ours 10k SFT & 85.70 & 69.15 & 88.33 & 73.33 & 81.84 & 58.96 & \textbf{85.29} & \textbf{67.15} \\
\addlinespace[2pt]
\textbf{Qwen3.5-9B} & 82.46 & 69.65 & 80.83 & 66.67 & 79.48 & 55.66 & 80.92 & 63.99 \\
\quad + WebArbiter 10k SFT & 85.57 & 68.16 & 90.83 & 73.33 & 77.12 & 52.36 & 84.51 & 64.62 \\
\rowcolor{gray!20}\quad + Ours 10k SFT & 87.44 & 70.65 & 88.33 & 73.33 & 83.14 & 60.85 & \textbf{86.30} & \textbf{68.28} \\
\bottomrule
\end{tabular}
\end{table}

\paragraph{Benchmarks and Evaluation Protocol.}
We evaluate comparative Web PRMs across two tasks:
(1) \textbf{\textsc{WebPRMBench}}: State-conditioned pairwise preference discrimination evaluated on Pairwise and Best-of-$N$ (BoN) Accuracy across \textsc{WebArena}~\citep{zhou2024webarena}, \textsc{AssistantBench}~\citep{yoran2024assistantbench}, and \textsc{WorkArena}~\citep{drouin2024workarena}. Because the demonstration trajectories used for SFT (in both \method and the WebArbiter baseline) are drawn from the WPRM demonstration pool~\citep{chae2025webshepherd} built on \textsc{Mind2Web} environments~\citep{deng2023mind2web}, the \textsc{Mind2Web} subset of \textsc{WebPRMBench} overlaps with the SFT training distribution. We therefore exclude it, so that evaluation on the three unseen environments directly measures the cross-scenario transferability of the trained PRMs rather than in-distribution memorization.
(2) \textbf{\textsc{WebArena-Lite}}~\citep{liu2024visualagentbench}: Reward-guided trajectory search across five web environments, evaluating whether comparative PRMs improve downstream agent performance. At each step, the policy generates five candidate actions, and the PRM selects one via a knockout tournament~\citep{guo2025reward}. We evaluate GPT-4o and GPT-4o-mini as policy model, reporting task success rates and absolute gains ($\boldsymbol{\Delta}$) over the corresponding policies without PRM-guided search.

\paragraph{Evaluated Models and Baselines.}
We benchmark: (1)~\textbf{Closed-Source LLMs}: GPT-5~\citep{openai2025gpt5}, Claude-4.6-Sonnet~\citep{anthropic2026claude46}, Gemini-3.7-Flash~\citep{googledeepmind2026gemini37flash}, and jev-1.13~\citep{typesafe2026jev113} (Appendix~\ref{sec:app_prompts_jev}); (2)~\textbf{Open-Source LLMs}: Llama-3-70B-Instruct~\citep{grattafiori2024llama}, DeepSeek-R1~\citep{deepseek2025r1}, and DeepSeek-V4-Flash-0731~\citep{deepseek2026v4}; and (3)~\textbf{Fine-Tuned Comparative Web PRMs}: Baselines \textsc{WebShepherd}-3B/8B~\citep{chae2025webshepherd}, alongside systematic comparisons across six open-source Qwen backbones (3B--9B)~\citep{qwen2024qwen25,qwen2025qwen3,qwen2026qwen35} fine-tuned on WebArbiter 10k SFT~\citep{zhang2026webarbiter} vs.\ \methoddata under an aligned $\sim$10k budget and identical protocols.

\subsection{Main Experimental Results}
\label{sec:main_results}

\paragraph{Step-Level Process Preference Discrimination on \textsc{WebPRMBench}.}
Table~\ref{tab:web_prm_bench} presents preference discrimination results on \textsc{WebPRMBench}:
(1) \textbf{Consistent Macro-Average Superiority Across Scales}: Models trained on \methoddata consistently outperform the WebArbiter baseline across all six backbones (3B to 9B) in both macro-average Pairwise and BoN metrics (e.g., Qwen2.5-3B achieves +6.45\% Pairwise and +7.76\% BoN gains; Qwen3.5-9B reaches 86.30\% / 68.28\%), yielding the highest macro-average discrimination at every scale.
(2) \textbf{Data Efficiency \& Generalization}: Qwen2.5-3B trained on \methoddata (80.65\% Pairwise) surpasses larger models trained on WebArbiter (Qwen2.5-7B at 80.58\%, Qwen3-8B at 80.45\%), peaking on enterprise \textsc{WorkArena} (Qwen3.5-4B: 81.84\% / 58.96\% vs.\ 76.89\% / 47.64\%) and \textsc{AssistantBench} (Qwen3-8B: 88.33\% / 73.33\% vs.\ 80.83\% / 60.00\%).
(3) \textbf{Competitive with Frontier LLMs}: Fine-tuned open-source PRMs match or exceed proprietary LLMs like Claude-4.6-Sonnet (82.53\% / 65.88\%) and GPT-5 (82.55\% / 66.53\%).

\paragraph{Downstream Reward-Guided Trajectory Search on \textsc{WebArena-Lite}.}
\begin{table}[t]
\centering
\caption{\textbf{Downstream Trajectory Search on WebArena-Lite.} Success rates (\%) across 5 environments with Best-of-$N$ search guided by Web PRMs. Avg.\ is macro-average; $\Delta$ is absolute gain over unguided execution. $*$ denotes baseline results directly reported from the \textsc{WebArbiter} paper. Rows trained on \methoddata are shaded; \textbf{bold} marks best within each policy group.}
\label{tab:downstream_search}
\footnotesize
\setlength{\tabcolsep}{3pt}
\renewcommand{\arraystretch}{1.0}
\begin{tabular}{llccccccc}
\toprule
\textbf{Policy Model} & \textbf{WebPRM Guidance} & \textbf{Shopping} & \textbf{CMS} & \textbf{Reddit} & \textbf{GitLab} & \textbf{MAP} & \textbf{Avg.} & $\boldsymbol{\Delta}$ \\
\midrule
\multirow{5}{*}{GPT-4o-mini}
& w/o Trajectory Search* & 21.74 & 22.86 & 19.05 & 34.38 & 19.35 & 23.48 & -- \\
& GPT-4o-mini* & 24.44 & 22.86 & 26.32 & 33.33 & 15.38 & 24.47 & +0.99 \\
& WebShepherd-8B* & 26.09 & 45.71 & 23.81 & 40.62 & 35.48 & 34.34 & +10.87 \\
& Qwen3.5-9B + WebArbiter 10k & 34.78 & 38.89 & 33.33 & 38.24 & 19.35 & 32.92 & +9.44 \\
\rowcolor{gray!20}\cellcolor{white} & Qwen3.5-9B + Ours 10k & 36.96 & 30.56 & 50.00 & 38.24 & 25.81 & \textbf{36.31} & \textbf{+12.83} \\
\midrule
\multirow{5}{*}{GPT-4o}
& w/o Trajectory Search* & 23.91 & 31.43 & 28.57 & 56.25 & 19.35 & 31.90 & -- \\
& GPT-4o-mini* & 26.67 & 37.14 & 42.11 & 40.00 & 19.23 & 33.03 & +1.13 \\
& WebShepherd-8B* & 30.43 & 42.86 & 47.62 & 46.88 & 35.48 & 40.65 & +8.75 \\
& Qwen3.5-9B + WebArbiter 10k & 34.78 & 30.56 & 45.83 & 44.12 & 35.48 & 38.15 & +6.25 \\
\rowcolor{gray!20}\cellcolor{white} & Qwen3.5-9B + Ours 10k & 47.83 & 50.00 & 62.50 & 41.18 & 29.03 & \textbf{46.11} & \textbf{+14.21} \\
\bottomrule
\end{tabular}
\end{table}

Table~\ref{tab:downstream_search} evaluates comparative Web PRMs as search guides over external policy generators on \textsc{WebArena-Lite}:
(1) \textbf{Macro-Average Guidance Gains}: Across environments, our PRM (\method, Qwen3.5-9B + Ours 10k) outperforms unguided execution, \textsc{WebShepherd}, and \textsc{WebArbiter}. Guiding GPT-4o, it achieves 46.11\% macro-average success (+14.21\% over unguided, +7.96\% over WebArbiter); guiding lightweight GPT-4o-mini, it reaches 36.31\% (+12.83\%), surpassing unguided GPT-4o (31.90\%).
(2) \textbf{Resolving High-Branching Bottlenecks}: \method excels in complex environments: under GPT-4o, it doubles success on Shopping (23.91\% $\to$ 47.83\%) and leads on Reddit (28.57\% $\to$ 62.50\%) and CMS (31.43\% $\to$ 50.00\%); under GPT-4o-mini, it lifts Reddit from 19.05\% to 50.00\%. On GitLab, guidance declines (56.25\% $\to$ 41.18\% under GPT-4o), likely because code repository UIs diverge from consumer web domains in training demonstrations.
(3) \textbf{Effective Plug-and-Play Verifier}: Unlike policy self-evaluation (+0.99\%), \method exhibits robust out-of-policy discrimination across distinct generators, confirming the practical utility of comparative process reward modeling.

\subsection{In-Depth Analysis and Ablation Studies}
\label{sec:ablation}

\begin{figure}[t]
\begin{minipage}[t]{0.49\textwidth}
\centering
\includegraphics[width=\linewidth,height=2.5cm,keepaspectratio]{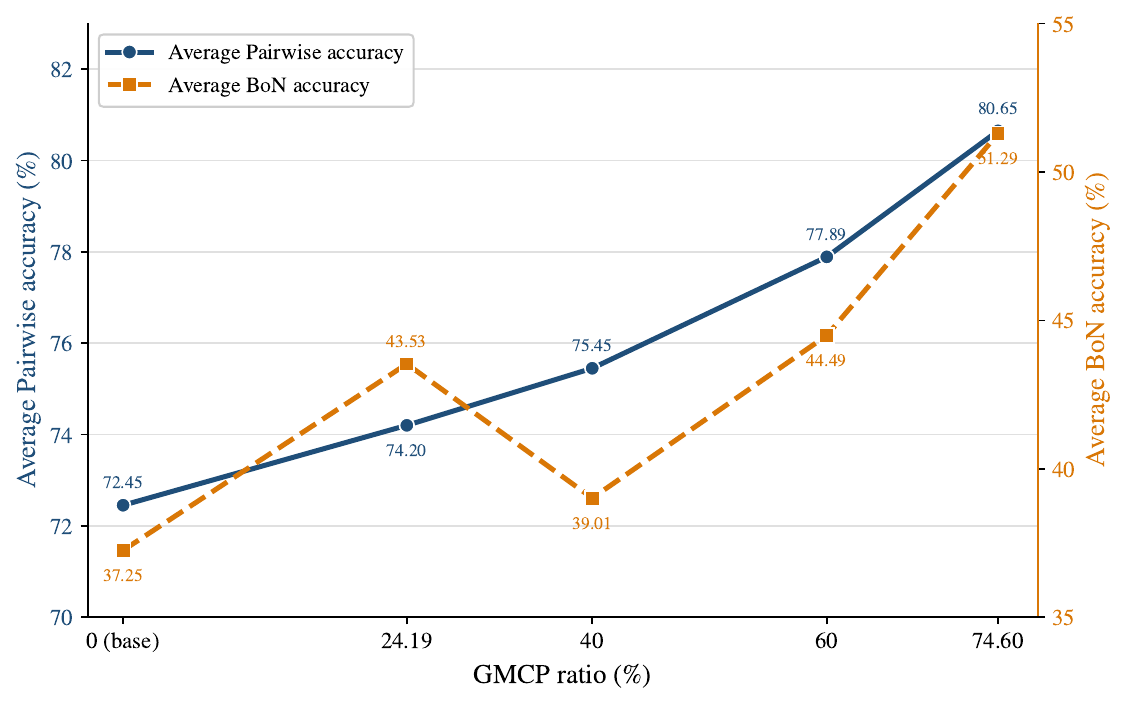}
\caption{\textbf{Effect of GMCP Ratio.} Average Pairwise/BoN accuracy of Qwen2.5-3B-Instruct under controlled GMCP ratios (24.19\%--74.60\%) vs.\ un-finetuned base (0\%); overall performance exhibits an upward scaling trend as the GMCP ratio increases (Table~\ref{tab:gmcp_ratio} in Appendix~\ref{sec:app_extended_eval}).}
\label{fig:gmcp_ratio}
\end{minipage}\hfill
\begin{minipage}[t]{0.49\textwidth}
\centering
\includegraphics[width=\linewidth,height=2.5cm,keepaspectratio]{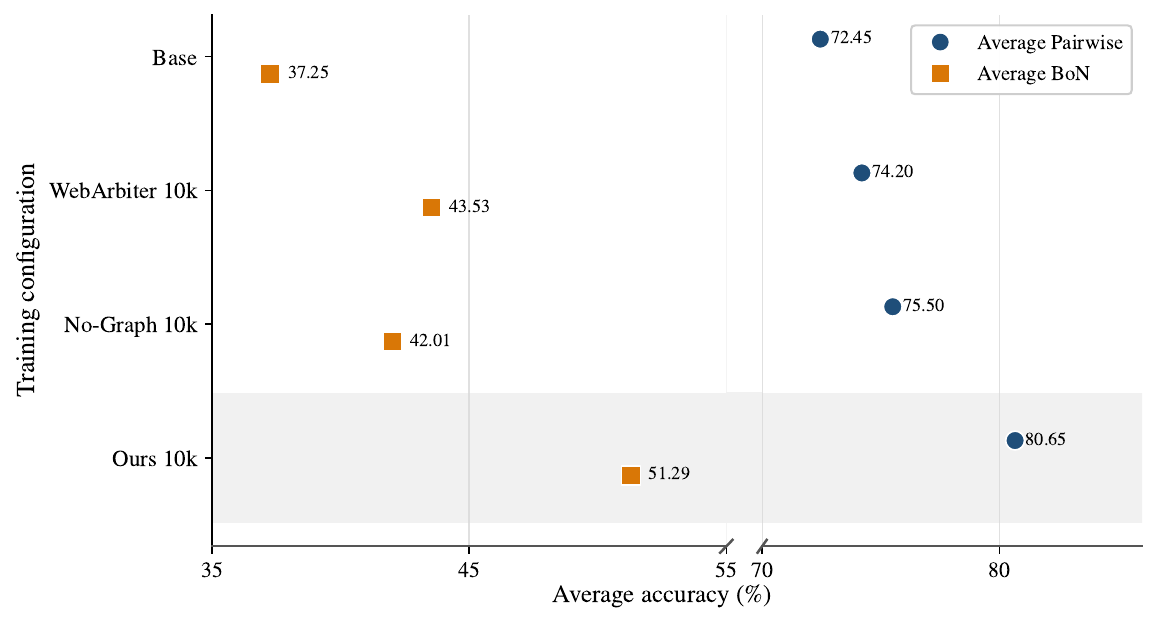}
\caption{\textbf{Ablation on Graph Necessity.} Average Pairwise/BoN accuracy of PRMs on WebArbiter, No-Graph, and \method (\methoddata); graph topology provides critical structural signals (Table~\ref{tab:no_graph_ablation} in Appendix~\ref{sec:app_extended_eval}).}
\label{fig:no_graph_ablation}
\end{minipage}
\end{figure}

\paragraph{Impact of the Grounded Minimal Contrastive Pair (GMCP) Ratio.}
We evaluate Qwen2.5-3B across controlled GMCP ratios under an aligned $\sim$10k budget (Figure~\ref{fig:gmcp_ratio}; Table~\ref{tab:gmcp_ratio} in Appendix~\ref{sec:app_extended_eval}): (1) \textbf{Overall Upward Scaling Trend}: Macro-average Pairwise Accuracy climbs from 74.20\% (24.19\% GMCP) to 80.65\% (74.60\%), and BoN accuracy improves from 43.53\% to 51.29\%, confirming that higher proportions of grounded minimal contrastive supervision yield superior discrimination. (2) \textbf{Contextual Decisions}: We additionally partition the pooled test pairs of all three benchmark environments by whether both candidates operate on genuine page elements with identical action types: on the \textit{Full GMCP subset}, BoN accuracy scales steadily: 45.89\% (base) $\to$ 66.67\% (WebArbiter) $\to$ 71.50\% (Ours, +4.83\% over WebArbiter; Pairwise reaching 85.71\%). Conversely, on the \textit{Full Non-GMCP subset}, accuracy saturates early across all variants ($\sim$51\% BoN), confirming that gains stem from contextual decision-making rather than superficial heuristics.

\paragraph{Ablation on Graph Necessity.}
We compare against a strictly \textbf{No-Graph} baseline (Figure~\ref{fig:no_graph_ablation}; Table~\ref{tab:no_graph_ablation}, Appendix~\ref{sec:app_extended_eval}) mining negatives purely from local AXTree snapshots without cross-trajectory graphs. Both settings share identical demonstration states, BID remapping, teacher scoring ($K=5$), matched GMCP ratios (75.56\% vs 74.60\%), and aligned scales (9,643 vs 9,821 pairs). (1) \textbf{Clear Graph Advantage}: Ours 10k SFT achieves 80.65\% Pairwise and 51.29\% BoN, substantially outperforming No-Graph 10k SFT (75.50\% / 42.01\%) and WebArbiter (74.20\% / 43.53\%), with large gains on multi-step benchmarks like AssistantBench (85.00\% / 56.67\% vs.\ 70.00\% / 30.00\%) and WorkArena (75.35\% / 41.98\% vs.\ 72.05\% / 36.32\%). (2) \textbf{Topological Signals}: This confirms that cross-trajectory topological signals provide critical structural discrimination that local heuristics cannot replicate.

\paragraph{Ablation on Graph Confusion Axes.}
\begin{wrapfigure}{r}{0.23\textwidth}
\vspace{-1.4em}
\centering
\includegraphics[width=\linewidth]{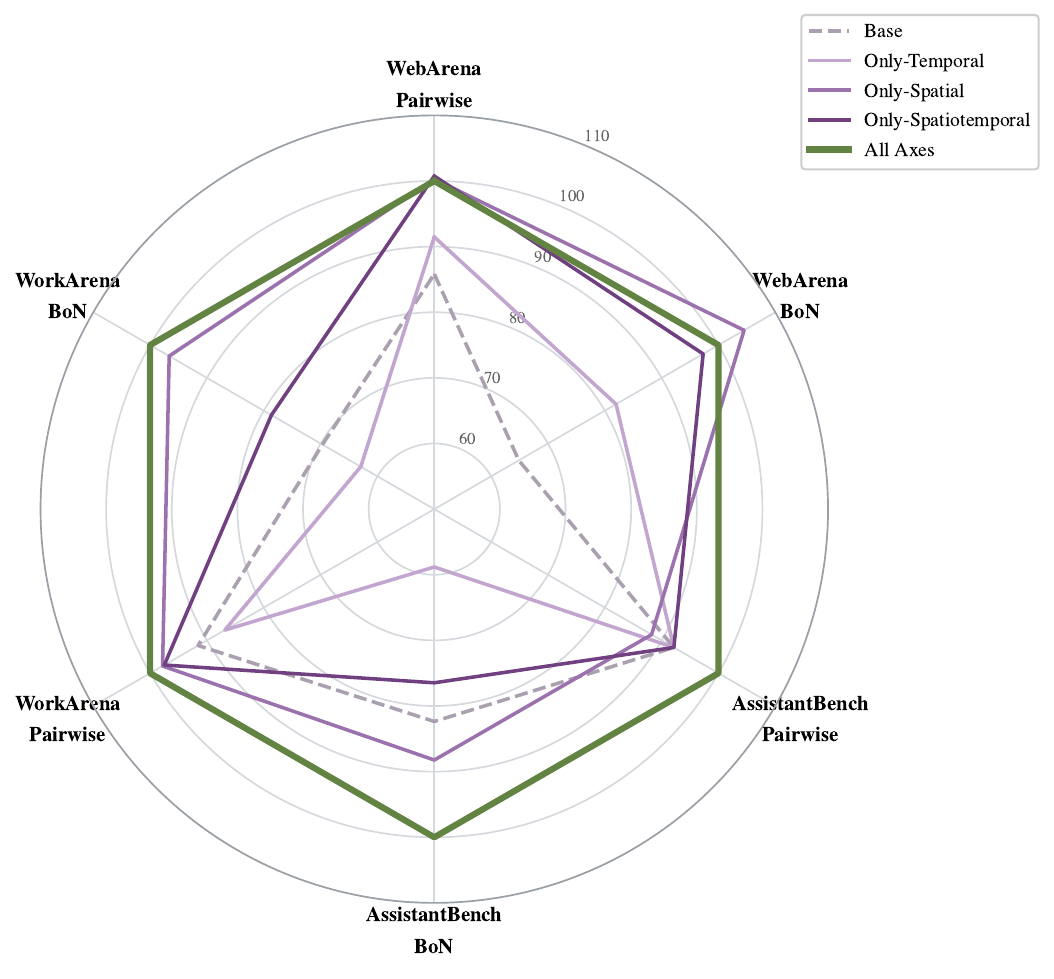}
\caption{Axis ablation.}
\label{fig:ablation_axes}
\vspace{-1.4em}
\end{wrapfigure}
Evaluating PRMs trained on individual axes (Figure~\ref{fig:ablation_axes}; Table~\ref{tab:ablation_axes} in Appendix~\ref{sec:app_extended_eval}) reveals: (1) \textbf{No Single Axis Is Sufficient}: Each axis exhibits clear weaknesses---\texttt{Only-Spatial} (76.85\% / 49.43\%) and \texttt{Only-Spatiotemporal} (78.00\% / 43.36\%) lag behind the full mixture, while \texttt{Only-Temporal} (74.63\% / 35.01\%) underperforms substantially on AssistantBench and WorkArena. (2) \textbf{All Three Axes Are Necessary}: The full multi-axis mixture achieves highest macro-average (80.65\% / 51.29\%), outperforming every single-axis variant on average Pairwise and BoN accuracy and confirming that the three axes are complementary and jointly necessary across disparate web environments (Figure~\ref{fig:ablation_axes}; \texttt{All Axes} normalized to 100\% to illustrate the relative performance).

\section{Conclusion}
\label{sec:conclusion}

We introduce \method, a graph-guided framework for synthesizing environment-grounded, minimally contrastive process preference supervision for Web PRMs. By proposing negatives across spatial, temporal, and spatiotemporal axes on an Interaction Element Graph, \method elevates the GMCP proportion from 24.19\% to 74.60\% in \methoddata, eliminating shallow shortcuts like element hallucinations. Across six backbones (3B--9B), PRMs trained on \methoddata outperform baselines on \textsc{WebPRMBench} and provide substantial search gains (+14.21\% on GPT-4o, +12.83\% on GPT-4o-mini) on \textsc{WebArena-Lite}.

% \newpage
\subsection*{Reproducibility Statement}
To facilitate the reproduction of our findings, we provide complete algorithmic, implementation, and experimental specifications across the paper.
The Interaction Element Graph formulation, negative proposal rules, and action mappings are detailed in Section~\ref{sec:method_graph} and Appendices~\ref{sec:app_rules} and \ref{sec:app_action_mapping}.
The data synthesis pipeline, including Action-Only formatting, teacher verification ($K=5$), quality assurance, and full prompt templates, is described in Sections~\ref{sec:method_teacher_synthesis} and \ref{sec:data_qa} and Appendices~\ref{sec:app_implementation}, \ref{sec:app_prompts_filtering}, and \ref{sec:app_qa_details}.
Evaluation setups and extended per-environment ablations are provided in Section~\ref{sec:exp_setup} and Appendix~\ref{sec:app_extended_eval}.
Our synthesized datasets and trained model checkpoints are anonymously accessible at \anonymousrepo, and full source code will be publicly released upon acceptance.

\subsection*{AI Use Statement}
\label{sec:app_ai_statement}

In this work, we used generative AI tools to generate synthetic datasets, implement methods, assist with translation, clean and reformat dataset, support qualitative and thematic data analysis, and interpret results. We have not used generative AI tools to propose or refine hypotheses, design or provide feedback on research methodology or experiments, help develop theoretical models or conceptual frameworks, formulate mathematical claims, provide critical ingredients for proving mathematical claims, or assist in the writing of proofs (not applicable to this work). Additionally, we used generative AI tools to create or edit software code and identify relevant literature. We have reviewed all AI-assisted work. LLM-generated code was verified and tested for correctness by authors. We take responsibility for the final content of this work, including text, claims, or artifacts produced with the aid of generative AI.

\bibliography{iclr2027_conference}
\bibliographystyle{iclr2027_conference}

\newpage
\appendix
\section{Formal Definitions of Graph-Based Proposal Rules}
\label{sec:app_rules}

Let $s_t = (I, O_t, H_t, u_t)$ denote the environmental state, $a_t^+ = (\mathrm{op}^+, v_t^+, \text{val}^+)$ the positive step targeting element $v_t^+ \in V$, and $\mathcal{V}_{\text{grounded}}(s_t)$ the set of elements in $G$ uniquely resolvable in $O_t$. 
The distance $d_\tau(u, v)$ denotes the shortest path distance from $u$ to $v$ along temporal edges $E_\tau$. 
The distance $d_s(u, v)$ denotes the shortest path distance from $u$ to $v$ along spatial edges $E_s$. 
Let $d_{\text{comb}}(u, v)$ denote the shortest graph distance between $u$ and $v$ on the combined element graph formed by spatial edges $E_s$ and temporal edges $E_\tau$. 
Let $\text{sim}_{\text{text}}(u, v) \in [0, 1]$ denote the character-level sequence-matching ratio between the normalized textual descriptors of $u$ and $v$.

For each proposed candidate node $v^- \in \mathcal{V}_{\text{grounded}}(s_t)$, the negative action is represented as $a^- = (\mathrm{op}^-, v^-, \text{val}^-)$. Rather than copying the positive step's operation or value, $\mathrm{op}^-$ and $\text{val}^-$ are instantiated by preferring the most frequently demonstrated complete action for that node across trajectories (falling back to the element's observed executable action label in the current observation, e.g., \texttt{click} for clickable roles or \texttt{fill} with placeholder values subsequently completed by an LLM, when demonstration history is absent). The six proposal rules $\mathcal{R} = \{r_1, \dots, r_6\}$ are formalized as follows:

\paragraph{1. Structurally Near.}
\begin{equation}
    \mathcal{N}_{\text{struct\_near}}(s_t, a_t^+) = \{ (\mathrm{op}^-, v^-, \text{val}^-) \mid v^- \in \mathcal{V}_{\text{grounded}}(s_t) \setminus \{v_t^+\}, \, 1 \le d_s(v_t^+, v^-) \le 5 \}
\end{equation}
where $d_s(v_t^+, v^-)$ is the path distance along spatial containment edges $E_s$.

\paragraph{2. Textually Near, Graph-Distant.}
\begin{equation}
\begin{aligned}
    \mathcal{N}_{\text{text\_near\_graph\_dist}}(s_t, a_t^+) = \{ (\mathrm{op}^-, v^-, \text{val}^-) \mid \, & v^- \in \mathcal{V}_{\text{grounded}}(s_t) \setminus \{v_t^+\}, \\
    & \text{sim}_{\text{text}}(v_t^+, v^-) \ge 0.75, \\
    & d_{\text{comb}}(v_t^+, v^-) > 4 \}
    % \lor d_{\text{comb}}(v_t^+, v^-) = \infty 
\end{aligned}
\end{equation}
Elements reachable within $\le 4$ hops on the combined element graph (containment plus interaction edges) across trajectories are excluded, as they often constitute valid multi-step navigation paths observed across trajectories.

\paragraph{3. Repeat Past.}
\begin{equation}
\begin{aligned}
    \mathcal{N}_{\text{repeat\_past}}(s_t, a_t^+) = \{ (\mathrm{op}^-, v_{t-k}^+, \text{val}^-) \mid \, & 1 \le k \le \min(N, t-1), \\
    & v_{t-k}^+ \in \mathcal{V}_{\text{grounded}}(s_t) \setminus \{v_t^+\}, \, \text{url}(v_{t-k}^+) = \text{url}(s_t) \}
\end{aligned}
\end{equation}
where $k$ indexes prior successful targeted interactions within the current trajectory (rather than raw environment steps) sharing the current page URL ($\text{url}(v_{t-k}^+) = \text{url}(s_t)$). The negative action is executed using the demonstrated action identity of node $v_{t-k}^+$. In our implementation, we set the lookback window to $N=5$.

\paragraph{4. Act Too Early.}
\begin{equation}
\begin{aligned}
    \mathcal{N}_{\text{act\_early}}(s_t, a_t^+) = \{ (\mathrm{op}^-, v_{t+k}^+, \text{val}^-) \mid \, & 1 \le k \le \min(N, T - t), \\
    & v_{t+k}^+ \in \mathcal{V}_{\text{grounded}}(s_t) \setminus \{v_t^+\}, \, \text{url}(v_{t+k}^+) = \text{url}(s_t) \}
\end{aligned}
\end{equation}
where $T$ denotes the total number of demonstrated steps in the trajectory, and $k$ indexes subsequent successful targeted interactions scheduled ahead in the demonstration trajectory that share the current page URL ($\text{url}(v_{t+k}^+) = \text{url}(s_t)$). The negative action is executed using the demonstrated action identity of node $v_{t+k}^+$. We set the lookahead horizon to $N=5$.

\paragraph{5. Away from Goal.}
Let $v_{\text{goal}}$ denote the final interaction element executed in the current task trajectory. When $d_{\text{comb}}(v_t^+, v_{\text{goal}}) < \infty$, the rule proposes:
\begin{equation}
\begin{aligned}
    \mathcal{N}_{\text{away\_goal}}(s_t, a_t^+) = \{ (\mathrm{op}^-, v^-, \text{val}^-) \mid \, & v^- \in \mathcal{V}_{\text{grounded}}(s_t) \setminus \{v_t^+\}, \\
    & d_{\text{comb}}(v^-, v_{\text{goal}}) > d_{\text{comb}}(v_t^+, v_{\text{goal}}) \}
\end{aligned}
\end{equation}
If $d_{\text{comb}}(v_t^+, v_{\text{goal}}) = \infty$ (i.e., the current positive target has no path to the goal element on the combined element graph), this rule is skipped ($\mathcal{N}_{\text{away\_goal}}(s_t, a_t^+) = \emptyset$).

\paragraph{6. Create a Cycle.}
\begin{equation}
\begin{aligned}
    \mathcal{N}_{\text{cycle}}(s_t, a_t^+) = \{ (\mathrm{op}^-, v^-, \text{val}^-) \mid \, & v^- \in \mathcal{V}_{\text{grounded}}(s_t) \setminus \{v_t^+\}, \\
    & \exists k \in \{1, \dots, \min(N, t-1)\}, \, d_\tau(v^-, v_{t-k}^+) < \infty \}
\end{aligned}
\end{equation}
with $N=5$. Here, candidate node $v^-$ need not have been visited previously itself; rather, transitioning to $v^-$ can reach one of the last $N=5$ demonstrated historical interaction targets $v_{t-k}^+$ along temporal interaction edges $E_\tau$ ($d_\tau(v^-, v_{t-k}^+) < \infty$), inducing an interaction cycle of length $d_\tau(v^-, v_{t-k}^+) + 1$ back to recent history.

\section{Persistent Node Identity, Action Mapping, and Virtual Nodes}
\label{sec:app_action_mapping}

\paragraph{Persistent Element Identity Function ($\phi$).}
Runtime Backend Node IDs (BIDs) in AXTrees are fundamentally ephemeral: they fluctuate across page reloads, dynamic AJAX updates, client-side re-renderings, and separate browser sessions. To track and align elements across distinct trajectories, we construct a persistent element identity mapping function $v = \phi(\text{site},\, \text{URL},\, \text{elem\_ctx}) \in V_{\text{elem}}$ grounded on BID-invariant structural context:
\begin{itemize}[leftmargin=*,itemsep=1pt,topsep=1pt]
    \item \textbf{Site and URL Scope}: $\text{site}$ identifies the target website domain, and $\text{URL}$ records the verbatim page URL observed in the demonstration. Scoping by normalized URL paths ensures that semantically distinct pages within the same site (e.g., checkout buttons on separate cart vs.\ order confirmation endpoints) instantiate distinct nodes rather than collapsing into false topological aliases.
    \item \textbf{Canonical Element Context ($\text{elem\_ctx}$)}: The element context is a deterministic serialization of the element's ID-free structural position within the AXTree snapshot, defined by a four-tuple:
    \begin{enumerate}[label=(\roman*),leftmargin=*,itemsep=1pt,topsep=1pt]
        \item \textit{Accessible Descriptor}: The element's accessibility role and accessible name (e.g., \texttt{role="combobox", name="From"}), capturing the semantic function intended by web accessibility standards.
        \item \textit{Ancestor Descriptor Chain}: The ordered sequence of accessibility roles and names along the path from the element upward to the document root (e.g., \texttt{root} $\to$ \texttt{main} $\to$ \texttt{region "Flight search"} $\to$ \texttt{group}), omitting non-operable, generic wrapper containers.
        \item \textit{Sibling Position}: The ordinal index of the element among sibling nodes sharing the identical accessibility role, disambiguating repetitive UI arrays (e.g., consecutive date cells within a monthly calendar grid).
        \item \textit{Subtree Signature}: A compact canonical signature reflecting the immediate subtree structure (child roles and text labels), distinguishing structurally distinct containers even when sharing ancestor paths.
    \end{enumerate}
\end{itemize}
This mapping ensures that when an element reappears in subsequent sessions or independent trajectories under a mutated runtime BID, it resolves deterministically to the same persistent node $v \in V_{\text{elem}}$.

\paragraph{Page-Scoped Virtual Node Formulation.}
For actions that do not operate on physical DOM elements—specifically global viewport manipulations (\texttt{scroll}), browser-level navigation (\texttt{go\_back}, \texttt{goto}), tab lifecycle actions (\texttt{new\_tab}, \texttt{tab\_close}), and terminal task outputs (\texttt{respond})—no concrete DOM BID exists. To represent such actions within the Interaction Element Graph $G = (V, E_s, E_\tau)$, we define typed Virtual Nodes $v_{\text{virt}} \in V_{\text{virt}}$ under the same key function $\phi(\text{site}, \text{URL}, \text{ctx})$:
\begin{itemize}[leftmargin=*,itemsep=1pt,topsep=1pt]
    \item \textbf{Landing Page Scoping}: The $\text{URL}$ parameter in $\phi$ is set to the \textit{landing URL}---defined as the page URL observed immediately after executing the action (or the current page for terminal response actions). The context $\text{ctx}$ pairs the action type with the landing page's accessibility root descriptor.
    \item \textbf{Preserving Local Topological Fidelity}: Instantiating virtual nodes per landing page rather than as site-wide singletons is critical: a \texttt{scroll} on the search results page and a \texttt{scroll} on the product details page represent fundamentally distinct state transitions with different downstream reachable elements. Co-locating virtual nodes with their respective landing pages ensures that directed temporal interaction edges $(u \xrightarrow{a_t} v_{\text{virt}})$ and $(v_{\text{virt}} \xrightarrow{a_{t+1}} w)$ preserve the precise local navigational flow of the website, yielding the unified node set $V = V_{\text{elem}} \cup V_{\text{virt}}$.
\end{itemize}

\paragraph{Action Modality to Graph Mapping.}
Browser interactions in web agents encompass both element-grounded interactions and global page-level commands. We define standard mappings between raw environment actions and the Interaction Element Graph $G = (V, E_s, E_\tau)$:
\begin{itemize}[leftmargin=*,itemsep=1pt,topsep=1pt]
    \item \textbf{Element-Grounded Actions}: Actions targeting specific DOM elements—including \texttt{click(bid)}, \texttt{hover(bid)}, \texttt{fill(bid, text)}, \texttt{select\_option(bid, opt)}, \texttt{press(bid, key)}, \texttt{focus(bid)}, \texttt{clear(bid)}, \texttt{dblclick(bid)}, and \texttt{drag\_and\_drop(...)}—are mapped to concrete persistent element nodes $v \in V_{\text{elem}}$ via our persistent identity function $\phi$.
    \item \textbf{Global Viewport and Navigation Actions}: Actions that operate without an element anchor—\texttt{scroll(x, y)} (offset-based), \texttt{goto(url)}, and \texttt{go\_back()}—are mapped to typed global virtual nodes ($\texttt{virtual:viewport}$ and $\texttt{virtual:browser\_navigation}$) $v_{\text{virt}}^{\text{scroll}}, v_{\text{virt}}^{\text{nav}} \in V_{\text{virt}}$.
    \item \textbf{Tab Operations}: Tab management operations—\texttt{new\_tab()}, \texttt{tab\_close()}, and \texttt{tab\_focus(idx)}—are assigned to a session-level tab virtual node ($\texttt{virtual:browser\_tabs}$) $v_{\text{virt}}^{\text{tab}} \in V_{\text{virt}}$.
    \item \textbf{Terminal Task Responses}: Terminal actions that communicate directly with the user, e.g., \texttt{respond(answer)} (implemented as \texttt{send\_msg\_to\_user}), are mapped to a user-facing virtual node ($\texttt{virtual:user}$) $v_{\text{virt}}^{\text{user}} \in V_{\text{virt}}$.
\end{itemize}

\paragraph{Virtual Node Representation.}
Virtual nodes ensure that every successfully resolved action in a demonstration trajectory maintains a topological presence in $G$.\footnote{A small fraction of successful actions (685 targets) cannot be resolved in the AXTree/SOM marker space and are therefore excluded from the graph rather than approximated; one \texttt{keyboard\_type} action was discarded, and eight \texttt{mouse\_click} targets were recovered through a manual BID audit table.} Transitions between concrete elements and virtual nodes (e.g., scrolling down to locate a product link, or navigating back to a search page) are seamlessly recorded as directed temporal edges in $E_\tau$, preserving complete trajectory flow.

\paragraph{Structural Decomposition of the Non-GMCP Residual.}
Grounded Minimal Contrastive Pairs (GMCPs) require two criteria to be fulfilled simultaneously: (i) both candidate target elements must be physically present and uniquely resolvable in the active observation snapshot $O_t$, and (ii) both candidate actions must share the exact same operation type ($\mathrm{op}^+ = \mathrm{op}^-$). As reported in Section~\ref{sec:dataset_curation_stats}, \methoddata achieves a 74.60\% GMCP proportion (7,326 out of 9,821 verified pairs). 

An audit of the remaining non-GMCP pairs confirms that 0\% fail due to target elements being absent from the active AXTree snapshot; all proposed negative targets resolve uniquely to active BIDs in $O_t$. Instead, the 25.40\% residual decomposes cleanly into two structural sources:
\begin{enumerate}[leftmargin=*,itemsep=1pt,topsep=1pt]
    \item \textbf{Virtual Positive Steps (16.6\%)}: In web navigation demonstrations, essential steps frequently involve global non-DOM operations (e.g., \texttt{scroll} to locate elements, \texttt{send\_msg\_to\_user} to return task answers, or \texttt{go\_back} to recover from dead ends). When such steps serve as the positive demonstration action $a_t^+$, their target corresponds to a virtual node $v_{\text{virt}} \in V_{\text{virt}}$, whereas the competing negative action $a_t^-$ is sampled from the active page's concrete DOM elements in $\mathcal{V}_{\text{grounded}}(s_t)$. This contrast represents a crucial high-level strategic dilemma (e.g., ``continue scrolling to find the target link'' vs.\ ``clicking an irrelevant distractor button on the visible viewport''), but inherently violates both identical target grounding and action-type matching.
    \item \textbf{Affordance-Preserving Action-Type Mismatches (9.1\%)}: For pairs where both positive and negative actions target genuine DOM elements in $O_t$, the negative candidate adopts the operation type that was historically demonstrated on that element during expert trajectories (or inferred from its AXTree role affordance, such as \texttt{fill} for text inputs). This affordance-aware assignment ensures that negative actions are executable and semantically meaningful rather than syntactically invalid (e.g., executing a \texttt{fill} on a non-editable link). Because proposal rules traverse topological, spatial, or temporal relations without coercing the negative candidate to adopt $\mathrm{op}^+$, an action-type mismatch arises whenever the selected distractor element possesses a different native affordance from the positive target. In the sampled 10,000 candidate set, the most frequent mismatch directions are $\texttt{fill} \to \texttt{click}$ (422 pairs), $\texttt{press} \to \texttt{click}$ (137 pairs), $\texttt{hover} \to \texttt{click}$ (136 pairs), $\texttt{select\_option} \to \texttt{click}$ (65 pairs), and $\texttt{click} \to \text{others}$ (110 pairs).
\end{enumerate}
Together, these two factors account for 100\% of the non-GMCP pairs (16.6\% + 9.1\% = 25.7\% in the 10,000 candidate pool, aligning with the 74.60\% verified training set), confirming that the sub-100\% GMCP ratio is an intentional design consequence of supporting global workflow steps and element affordances rather than grounding failure.

\section{Implementation Details}
\label{sec:app_implementation}

\paragraph{Action-Only Input Isolation.}
\label{sec:app_action_only}
As part of the concrete implementation of the \method data construction pipeline, we enforce an \textit{Action-Only} input protocol across filtering, supervision synthesis, and PRM training. We first distinguish the actor's exploratory thought before generating an action (\textit{candidate rationale} $r_t$) from the evaluator's comparative reasoning in hindsight (\textit{judgment rationale} $j_t$). Concatenating candidate rationales into PRM inputs, as in prior pipelines (e.g., \citet{zhang2026webarbiter}), allows models to exploit generator-specific stylistic signatures and confidence discrepancies instead of contextual decision semantics. Under our protocol, all candidate rationales are stripped away: the teacher model and the downstream PRM receive solely $(s_t, a_t^1, a_t^2)$, forcing them to generate autonomous judgment rationales grounded purely in page affordances and task intent. Concretely, each training tuple is formatted as $d_t = (s_t, a_t^1, a_t^2, j_t^*, y_t^*)$, where $s_t$ contains only the instruction, the AXTree observation, the action history, and the URL, and $a_t^1, a_t^2$ are rendered as raw action descriptions (operation type, target BID, and arguments) without any accompanying candidate rationale.

\paragraph{Teacher Verification Cost.}
\label{sec:app_teacher_cost}
All teacher verification and gold supervision synthesis calls use \texttt{gpt-5.6-sol}. Table~\ref{tab:app_teacher_cost} summarizes the cost of verifying the 10,000 sampled candidate pairs. In total, 11,210 API calls were issued (1.12 calls per pair on average): 9,528 pairs (95.3\%) passed on the first attempt, 293 passed after one to three retries (168 on the second, 67 on the third, 40 on the fourth, and 18 on the fifth attempt), and 179 pairs failed all five attempts and were discarded as suspected false negatives, consistent with the 98.21\% agreement pass rate reported in Section~\ref{sec:dataset_curation_stats}. The run consumed 35.6M input tokens ($\sim$3.1k per call) and approximately 5.3M billed output tokens ($\sim$472 per call), corresponding to an average cost of \$0.0334 per pair (\$0.0299 per call). Combined with contextual value completion for \texttt{fill} (199 calls, $<$ \$1), the total teacher-side production cost of \methoddata amounts to approximately \$335.

\begin{table}[h]
\caption{Teacher Verification and Gold Supervision Synthesis Cost for the 10,000 Sampled Candidate Pairs.}
\label{tab:app_teacher_cost}
\centering
\small
\begin{tabular}{lr}
\toprule
\textbf{Metric} & \textbf{Value} \\
\midrule
Judged pairs & 10,000 \\
API calls & 11,210 \\
First-attempt pass & 9,528 (95.3\%) \\
Passed after retries & 293 \\
\quad 2nd / 3rd / 4th / 5th attempt & 168 / 67 / 40 / 18 \\
Exhausted all 5 attempts (filtered) & 179 \\
Average calls per pair & 1.12 \\
Input tokens & 35.6M ($\sim$3.1k per call) \\
Output tokens (billed) & $\sim$5.3M ($\sim$472 per call) \\
Average cost per pair & \$0.0334 \\
Average cost per call & \$0.0299 \\
\bottomrule
\end{tabular}
\end{table}

\section{Prompt Specifications: Alignment Filtering, Gold Synthesis, Fill Value Completion, and Jev Evaluation}
\label{sec:app_prompts_filtering}
\label{sec:app_prompts_gold}

\subsection{Alignment Filtering Prompt Specification}

To ensure that the Teacher Model operates strictly under the \method protocol
during alignment filtering, we adopt the WebArbiter-style pairwise evaluation
template shown below. The two candidates are
presented as \emph{Response 1} and \emph{Response 2}, whose order is randomized
per pair (fair coin), with the true position of the grounded-correct action
recorded as the ground-truth label $y_t^* \in \{\text{response\_1},
\text{response\_2}\}$.

\begin{quote}
\small
\textbf{[System]} \\
You are a skilled expert at evaluating assistant responses. You should evaluate
given responses based on the judging criteria. Given the context of the
conversation and two responses from the Assistant, you need to refer to
determine the better response. Provide an overall comprehensive comparison upon
them.

\textbf{[User Input]} \\
\texttt{\#\#\#\#\ Intent\ \#\#\#\#} \texttt{\{instruction\}} \\
\texttt{\#\#\#\#\ AXTREE\ \#\#\#\#} Note: [bid] is the unique alpha-numeric identifier at
the beginning of lines for each element in the AXTree. Always use bid to refer
to elements in your actions. \texttt{\{accessibility\_tree\}} \\
\texttt{\#\#\#\#\ Trajectory\ \#\#\#\#} Note: The trajectory contains the sequence of
previous actions performed by the agent. \texttt{\{action\_history\}} \\
\texttt{\#\#\#\#\ start\ url\ \#\#\#\#} \texttt{\{start\_url\}} \\
\texttt{\#\#\#\#\ current\ url\ \#\#\#\#} The URL provides clues about the user's position
in the application flow. Use both the path and query parameters to infer page
type (e.g., homepage, search results, product detail, cart, checkout).
\texttt{\{current\_url\}} \\
\texttt{\#\#\#\#\ Assistant\ Responses\ \#\#\#\#} \\
\texttt{[The\ Begin\ of\ Response\ 1]} \\
\texttt{ACTION:\ \{action\_1\}} \\
\texttt{[The\ End\ of\ Response\ 1]} \\
\texttt{[The\ Begin\ of\ Response\ 2]} \\
\texttt{ACTION:\ \{action\_2\}} \\
\texttt{[The\ End\ of\ Response\ 2]} \\

\texttt{\#\#\#\ Output\ Instructions\ \#\#\#} \\
Format your output strictly using the following XML-style tags: \\
\texttt{<State>Summarize the current state based on the URL, AXTree, and
previous actions. Include what page the user is currently on, and what relevant
UI elements or information are visible.</State>} \\
\texttt{<Criteria>Other potential criteria specific to the query and the
context, and the weights of each criteria.</Criteria>} \\
\texttt{<Analysis>Compare Response 1 and Response 2 in detail according to the
<State> and <Criteria>.</Analysis>} \\
\texttt{<Answer>Response 1 or Response 2</Answer>} \\
Rules for \texttt{<Answer>}: \\
- If Response 1 is better, output exactly: \texttt{<Answer>Response 1</Answer>} \\
- If Response 2 is better, output exactly: \texttt{<Answer>Response 2</Answer>} \\
Important Notes: \\
- Be objective and base your evaluation strictly on the content of the
responses. \\
- Do not let the response order, length bias your judgment.
\end{quote}

\noindent The teacher decodes with temperature $0.2$ under a fixed per-attempt
seed. Given the extracted verdict $\text{Verdict} \in \{\text{Response 1},
\text{Response 2}\}$, filtering proceeds in up to $K{=}5$ \emph{sequential}
re-evaluation attempts.
% The first attempt is a direct reply; on each mismatch the same prompt is re-sent with a neutral nudge appended --- \emph{``Your previous choice did not match the correct evaluation. Please re-examine the intent, the page state (AXTree, URLs, trajectory) and both candidate actions, and produce a corrected \texttt{<State>}, \texttt{<Criteria>}, \texttt{<Analysis>} and \texttt{<Answer>} using the same format. Be objective and precise.''}  --- without ever revealing $y_t^*$, and the loop stops at the first aligned verdict.
A pair is discarded if and only if the verdict still disagrees with $y_t^*$ after all attempts.

\subsection{Gold Judgment Rationale and Verdict Synthesis}

For pairs that pass the alignment filtering, the assistant response from the
\emph{first} attempt whose verdict coincides with $y_t^*$ is retained as the
gold response; no further selection is applied among multiple aligned trials.
The retained gold text naturally decomposes into a reasoning prefix
$j_t^* = \langle\texttt{State}\rangle \circ \langle\texttt{Criteria}\rangle
\circ \langle\texttt{Analysis}\rangle$ followed by the verdict token
$y_t^* = \langle\texttt{Answer}\rangle$, so the final supervision target is
\begin{equation}
    Y_t = j_t^* \circ y_t^*,
\end{equation}
which is inserted verbatim as the assistant message of the SFT triple
$($system, user, assistant$)$.

\subsection{Contextual Fill Value Completion Prompt Specification}
\label{sec:app_prompts_fill}

As outlined in Section~\ref{sec:data_qa} and Appendix~\ref{sec:app_qa_details}, when candidate negative proposals select a text-entry element via graph mining that was not interacted with during the demonstration trajectory, the negative action lacks a concrete string argument. Rather than inserting trivial dummy text or risking format hallucinations, we prompt an LLM to generate a concise, syntactically valid, and contextually plausible input value that differs from any existing content in the field. The completion prompt consists of a fixed system instruction and a state-conditioned user prompt:

\begin{quote}
\small
\textbf{[System]} \\
You complete the text value for a fixed browser fill action. The target BID and action type are immutable. Return only a JSON object with one string field named value. Choose a concise, non-empty value that is syntactically valid and locally plausible for the target input in the current page state. The value must differ from every current value listed in the request. Do not output an action, BID, explanation, Markdown, or extra fields.

\textbf{[User Input]} \\
\texttt{User\ instruction:} \\
\texttt{\{instruction\}} \\
\texttt{Current\ URL:} \\
\texttt{\{current\_url\}} \\
\texttt{Previous\ actions:} \\
\texttt{\{action\_history\}} \\
\texttt{Fixed\ negative\ target\ BID:\ \{target\_id\}} \\
\texttt{Target\ description:\ \{target\_name\}} \\
\texttt{Current\ AXTree\ evidence:} \\
\texttt{\{target\_evidence\}} \\
\texttt{Values\ already\ present\ in\ the\ target\ (do\ not\ repeat):} \\
\texttt{\{current\_values\}} \\
\texttt{Known\ positive\ action\ (do\ not\ copy\ its\ BID):} \\
\texttt{\{positive\_action\}} \\
\texttt{Supply\ only\ the\ value\ for\ fill(<fixed\ BID>,\ value).}
\end{quote}

\noindent The completion model decodes with temperature $0.2$.

\subsection{Jev Evaluation Prompt Specifications}
\label{sec:app_prompts_jev}

For the proprietary decision model \texttt{jev-1.13}~\citep{typesafe2026jev113} evaluated in Table~\ref{tab:web_prm_bench}, inputs are structured into typed state blocks (\texttt{State}) and decision queries (\texttt{Question}) according to the Jev API specification. We evaluate two decision modes: \textbf{Choice mode} and \textbf{Noul mode}. 

In both modes, the input context consists of two shared state blocks: \texttt{[State: judging\_guidance]} providing overall evaluation directives, and \texttt{[State: web\_agent\_context]} providing task intent, accessibility tree observation, trajectory history, URLs, and the two candidate assistant responses.

\begin{quote}
\small
\textbf{[State: judging\_guidance]} \\
You are a skilled expert at evaluating assistant responses. You should evaluate given responses based on the judging criteria. \\
Given the context of the conversation and two responses from the Assistant, you need to refer to determine the better response. Provide an overall comprehensive comparison upon them. \\
Be objective and base the evaluation strictly on the content of the two responses. Do not let the response order or response length bias the judgment.
\vspace{2mm}

\textbf{[State: web\_agent\_context]} \\
\texttt{\#\#\#\#\ Intent\ \#\#\#\#} \\
\texttt{\{intent\}} \\
\texttt{\#\#\#\#\ AXTREE\ \#\#\#\#} \\
Note: [bid] is the unique alpha-numeric identifier at the beginning of lines for each element in the AXTree. Always use bid to refer to elements in your actions. \\
\texttt{\{axtree\}} \\
\texttt{\#\#\#\#\ Trajectory\ \#\#\#\#} \\
Note: The trajectory contains the sequence of previous actions performed by the agent. \\
\texttt{\{trajectory\}} \\
\texttt{\#\#\#\#\ start\ url\ \#\#\#\#} \\
\texttt{\{start\_url\}} \\
\texttt{\#\#\#\#\ current\ url\ \#\#\#\#} \\
The URL provides clues about the user's position in the application flow. Use both the path and query parameters to infer page type (e.g., homepage, search results, product detail, cart, checkout). \\
\texttt{\{current\_url\}} \\
\texttt{\#\#\#\#\ Assistant\ Responses\ \#\#\#\#} \\
\texttt{[The\ Begin\ of\ Response\ 1]} \\
\texttt{\{response\_1\}} \\
\texttt{[The\ End\ of\ Response\ 1]} \\
\texttt{[The\ Begin\ of\ Response\ 2]} \\
\texttt{\{response\_2\}} \\
\texttt{[The\ End\ of\ Response\ 2]}
\end{quote}

Under \textbf{Choice Mode}, the question block is specified as:

\begin{quote}
\small
\textbf{[Question: better\_response]} \\
\texttt{type:\ choice} \\
\texttt{instructions:} Compare the assistant's Response 1 and Response 2 found in \texttt{state.web\_agent\_context} and decide which one is the better next step for the web agent. Follow the judging guidance in \texttt{state.judging\_guidance}, and check each response against the intent, the AXTree of the current page, the trajectory of previous actions, and the start/current URLs. Prefer the response whose thought and action are more correct, better grounded in the current page state, and more effective at progressing toward the intent. \\
\texttt{criteria:} \\
\texttt{Response 1:} Response 1, the block between \texttt{'[The Begin of Response 1]'} and \texttt{'[The End of Response 1]'}, is the better response: compared with Response 2 it is more correct, better grounded in the current page state, and more effective for the intent. \\
\texttt{Response 2:} Response 2, the block between \texttt{'[The Begin of Response 2]'} and \texttt{'[The End of Response 2]'}, is the better response: compared with Response 1 it is more correct, better grounded in the current page state, and more effective for the intent.
\end{quote}

Under \textbf{Noul Mode}, the two \texttt{[State]} blocks remain identical, and only the \texttt{[Question]} block is replaced with:

\begin{quote}
\small
\textbf{[Question: better\_response]} \\
\texttt{type:\ noul} \\
\texttt{instructions:} Comparing the assistant's Response 1 and Response 2 found in \texttt{state.web\_agent\_context} under the judging guidance in \texttt{state.judging\_guidance}, is Response 1 the better next step for the web agent? \\
\texttt{criteria:} \\
\texttt{true:} Response 1 is the better response: it is more correct, better grounded in the current page state, and more effective for the intent than Response 2. \\
\texttt{false:} Response 2 is the better response, or the two responses are indistinguishable in quality.
\end{quote}

\section{Extended Evaluation and Repeated Runs}
\label{sec:app_extended_eval}

\paragraph{Ablation on Graph Confusion Axes.}
Table~\ref{tab:ablation_axes} details the full benchmark performance of comparative Web PRMs fine-tuned on individual confusion axes (Spatial, Temporal, and Spatiotemporal) versus the complete multi-axis mixture using Qwen2.5-3B-Instruct. Because the temporal axis naturally yields fewer negative candidates from expert demonstration trajectories (470 pairs vs.\ $>$4k pairs for spatial and spatiotemporal axes; Table~\ref{tab:data_distributions}), we scale up its fine-tuning epochs so that the cumulative training compute and gradient update steps are aligned to the same order of magnitude across all single-axis configurations.

\begin{table}[t]
\centering
\caption{\textbf{Ablation on Graph Confusion Axes.} Performance comparison of comparative Web PRMs fine-tuned on individual graph confusion axes versus the complete multi-axis mixture (evaluated on \textsc{WebPRMBench} with Qwen2.5-3B-Instruct). Pairwise accuracy and state-level Best-of-$N$ (BoN) accuracy are reported in percent (\%). The row corresponding to our full confusion mixture is shaded, and \textbf{bold} indicates the best result in the Average columns.}
\label{tab:ablation_axes}
\small
\setlength{\tabcolsep}{3.5pt}
\renewcommand{\arraystretch}{1.05}
\resizebox{\textwidth}{!}{%
\begin{tabular}{lcccccccc}
\toprule
\multirow{2}{*}{\textbf{Model / Training Configuration}} & \multicolumn{2}{c}{\textbf{WebArena}} & \multicolumn{2}{c}{\textbf{AssistantBench}} & \multicolumn{2}{c}{\textbf{WorkArena}} & \multicolumn{2}{c}{\textbf{Average}} \\
\cmidrule(lr){2-3} \cmidrule(lr){4-5} \cmidrule(lr){6-7} \cmidrule(lr){8-9}
& Pairwise & BoN & Pairwise & BoN & Pairwise & BoN & Pairwise & BoN \\
\midrule
\textbf{Qwen2.5-3B-Instruct} & 70.02 & 35.82 & 78.33 & 46.67 & 68.99 & 29.25 & 72.45 & 37.25 \\
\quad + Only-Temporal& 74.63 & 45.27 & 78.33 & 33.33 & 65.45 & 26.42 & 72.80 & 35.01 \\
\quad + Only-Spatial& 81.84 & 57.71 & 75.00 & 50.00 & 73.70 & 40.57 & 76.85 & 49.43 \\
\quad + Only-Spatiotemporal& 82.21 & 53.73 & 78.33 & 43.33 & 73.47 & 33.02 & 78.00 & 43.36 \\
\rowcolor{gray!20}\quad + All Axes & 81.59 & 55.22 & 85.00 & 56.67 & 75.35 & 41.98 & \textbf{80.65} & \textbf{51.29} \\
\bottomrule
\end{tabular}%
}
\end{table}

\paragraph{Effect of the Grounded Minimal Contrastive Pair (GMCP) Ratio.}
Table~\ref{tab:gmcp_ratio} details the full benchmark performance of the GMCP-ratio ablation discussed in Section~\ref{sec:ablation}, comparing Qwen2.5-3B-Instruct fine-tuned on $\sim$10k SFT pairs with controlled GMCP ratios against the un-finetuned base model and the WebArbiter baseline.

\begin{table}[t]
\centering
\caption{\textbf{Effect of the Grounded Minimal Contrastive Pair (GMCP) Ratio.} Qwen2.5-3B-Instruct results with approximately 10k SFT pairs per fine-tuned variant. Pairwise and state-level Best-of-$N$ (BoN) accuracies are reported in percent (\%). The best results in the Average column are in \textbf{bold}.}
\label{tab:gmcp_ratio}
\small
\setlength{\tabcolsep}{2.5pt}
\renewcommand{\arraystretch}{1.05}
\resizebox{\textwidth}{!}{%
\begin{tabular}{lccccccccc}
\toprule
\multirow{2}{*}{\textbf{Model / SFT Data}} & \multirow{2}{*}{\textbf{GMCP Ratio}} & \multicolumn{2}{c}{\textbf{WebArena}} & \multicolumn{2}{c}{\textbf{AssistantBench}} & \multicolumn{2}{c}{\textbf{WorkArena}} & \multicolumn{2}{c}{\textbf{Average}} \\
\cmidrule(lr){3-4} \cmidrule(lr){5-6} \cmidrule(lr){7-8} \cmidrule(lr){9-10}
& & Pairwise & BoN & Pairwise & BoN & Pairwise & BoN & Pairwise & BoN \\
\midrule
\textbf{Qwen2.5-3B-Instruct} & -- & 70.02 & 35.82 & 78.33 & 46.67 & 68.99 & 29.25 & 72.45 & 37.25 \\
\quad + WebArbiter 10k & 24.19\% & 81.84 & 55.22 & 69.17 & 40.00 & 71.58 & 35.38 & 74.20 & 43.53 \\
\quad + GMCP40 10k & 40.00\% & 81.47 & 50.75 & 70.83 & 26.67 & 74.06 & 39.62 & 75.45 & 39.01 \\
\quad + GMCP60 10k & 60.00\% & 82.71 & 56.72 & 76.67 & 36.67 & 74.29 & 40.09 & 77.89 & 44.49 \\
\rowcolor{gray!20}\quad + Ours 10k & 74.60\% & 81.59 & 55.22 & 85.00 & 56.67 & 75.35 & 41.98 & \textbf{80.65} & \textbf{51.29} \\
\bottomrule
\end{tabular}%
}
\end{table}

\paragraph{Graph Necessity Ablation.}
Table~\ref{tab:no_graph_ablation} details the full benchmark performance of the Graph Necessity ablation discussed in Section~\ref{sec:ablation}, comparing \method against the No-Graph variant and the WebArbiter baseline.

\begin{table}[t]
\centering
\caption{\textbf{Ablation on Graph Necessity.} Comparison of comparative Web PRMs trained with the WebArbiter baseline, the No-Graph variant, and \method on \textsc{WebPRMBench}. Pairwise accuracy and state-level Best-of-$N$ (BoN) accuracy are reported in percent (\%). The best results in the Average columns are in \textbf{bold}.}
\label{tab:no_graph_ablation}
\small
\setlength{\tabcolsep}{3.5pt}
\renewcommand{\arraystretch}{1.05}
\begin{tabular}{lcccccccc}
\toprule
\multirow{2}{*}{\textbf{Model / Training Configuration}} & \multicolumn{2}{c}{\textbf{WebArena}} & \multicolumn{2}{c}{\textbf{AssistantBench}} & \multicolumn{2}{c}{\textbf{WorkArena}} & \multicolumn{2}{c}{\textbf{Average}} \\
\cmidrule(lr){2-3} \cmidrule(lr){4-5} \cmidrule(lr){6-7} \cmidrule(lr){8-9}
& Pairwise & BoN & Pairwise & BoN & Pairwise & BoN & Pairwise & BoN \\
\midrule
\textbf{Qwen2.5-3B-Instruct} & 70.02 & 35.82 & 78.33 & 46.67 & 68.99 & 29.25 & 72.45 & 37.25 \\
\quad + WebArbiter 10k SFT & 81.84 & 55.22 & 69.17 & 40.00 & 71.58 & 35.38 & 74.20 & 43.53 \\
\quad + No-Graph 10k SFT & 84.45 & 59.70 & 70.00 & 30.00 & 72.05 & 36.32 & 75.50 & 42.01 \\
\rowcolor{gray!20}\quad + Ours 10k SFT & 81.59 & 55.22 & 85.00 & 56.67 & 75.35 & 41.98 & \textbf{80.65} & \textbf{51.29} \\
\bottomrule
\end{tabular}
\end{table}

\paragraph{Repeated Evaluation Runs and Statistical Stability.}
Table~\ref{tab:repeated_runs} reports the pairwise accuracy and state-level Best-of-$N$ (BoN) accuracy across 3 repeated evaluation runs on \textsc{WebPRMBench} using Qwen3.5-9B. 

\begin{table}[htbp]
\centering
\caption{\textbf{Multiple Repeated Evaluation Results on \textsc{WebPRMBench}.} Performance of comparative Web PRMs across repeated evaluation across 3 runs. Pairwise Accuracy (\%) and State-level Best-of-$N$ (BoN) Accuracy (\%) are reported as $\text{mean} \pm \text{std}$. Rows fine-tuned on our synthesized data (\methoddata) are shaded, and \textbf{bold} indicates the best mean result in the Average columns.}
\label{tab:repeated_runs}
\small
\setlength{\tabcolsep}{3.0pt}
\renewcommand{\arraystretch}{1.1}
\resizebox{\textwidth}{!}{%
\begin{tabular}{lcccccccc}
\toprule
\multirow{2}{*}{\textbf{Model}} & \multicolumn{2}{c}{\textbf{WebArena}} & \multicolumn{2}{c}{\textbf{AssistantBench}} & \multicolumn{2}{c}{\textbf{WorkArena}} & \multicolumn{2}{c}{\textbf{Average}} \\
\cmidrule(lr){2-3} \cmidrule(lr){4-5} \cmidrule(lr){6-7} \cmidrule(lr){8-9}
& Pairwise & BoN & Pairwise & BoN & Pairwise & BoN & Pairwise & BoN \\
\midrule
\textbf{Qwen3.5-9B} & 81.88 $\pm$ 0.50 & 68.32 $\pm$ 1.25 & 81.67 $\pm$ 0.84 & 66.67 $\pm$ 3.34 & 78.69 $\pm$ 0.68 & 54.72 $\pm$ 0.95 & 80.75 $\pm$ 0.21 & 63.24 $\pm$ 1.58 \\
\quad + WebArbiter 10k SFT & 85.28 $\pm$ 0.36 & 68.33 $\pm$ 0.29 & 89.72 $\pm$ 0.96 & 71.11 $\pm$ 1.92 & 77.48 $\pm$ 0.20 & 52.83 $\pm$ 0.81 & 84.16 $\pm$ 0.51 & 64.09 $\pm$ 1.01 \\
\rowcolor{gray!20}\quad + Ours 10k SFT & 87.31 $\pm$ 0.57 & 70.98 $\pm$ 1.04 & 87.78 $\pm$ 1.74 & 71.11 $\pm$ 3.85 & 83.57 $\pm$ 0.65 & 61.95 $\pm$ 2.76 & \textbf{86.22} $\pm$ 0.20 & \textbf{68.01} $\pm$ 0.24 \\
\bottomrule
\end{tabular}%
}
\end{table}

The results show low variance across runs (average Pairwise standard deviation is $\pm 0.20\%$, and average BoN standard deviation is $\pm 0.24\%$). On overall macro-average, our model fine-tuned on \methoddata demonstrates consistent improvements over both the un-finetuned base model and the WebArbiter 10k baseline (reaching $86.22 \pm 0.20\%$ Pairwise and $68.01 \pm 0.24\%$ BoN, compared to $84.16 \pm 0.51\%$ and $64.09 \pm 1.01\%$), with clear gains on WebArena and WorkArena while maintaining competitive performance on AssistantBench, confirming the statistical robustness of our gains.

\section{Data Quality Assurance Pipeline Details}
\label{sec:app_qa_details}

\paragraph{Parameter Completion for Fill Actions.}
Among the 262 \texttt{fill} negatives produced by graph traversal in the mined pool, 199 lacked a concrete input string value, since the candidate element was not typed into during the reference trajectory under identical context. To ensure environmental executability, we prompt an LLM with the page context and target input box attributes to generate a contextually plausible but incorrect or distracting input string (e.g., typing an incompatible postal code format or an alternative movie name) for each empty value (the full prompt specification and decoding protocol are detailed in Appendix~\ref{sec:app_prompts_fill}). In addition, 11 non-editable targets were normalized to \texttt{click} (Appendix~\ref{sec:app_qa_details}, Combobox Control Correction), yielding 251 \texttt{fill} negatives in the pool. The sampled set of 10,000 pairs contains 73 \texttt{fill} negatives, of which 68 survive in the final verified set of 9,821 pairs. Quality audits on the final set confirmed 0 empty values, 0 repetitions of current element values, and 0 invalid string formats.

\paragraph{Combobox Control Correction.}
During initial auditing, we observed that certain custom dropdown elements (e.g., ARIA role \texttt{combobox} without an underlying editable \texttt{<input>} tag) were assigned \texttt{fill} operations by heuristics. Attempting to execute \texttt{fill} on such elements causes runtime execution errors in browser environments. We audited all such controls and remapped their actions to \texttt{click}, resolving the control selection via standard dropdown click-and-select semantics.

% \section{Limitations and Future Work}
% \label{sec:app_limitations}

% \paragraph{Action Modality Imbalance.}
% In our final \methoddata dataset of 9,821 pairs, \texttt{click} operations account for 97.54\% of negative actions, while \texttt{fill} accounts for 0.69\%. Although this reflects the natural distribution of web interactions—where navigation, link following, tab selection, and filtering overwhelmingly involve click events—future work should explore targeted graph mining for complex multi-field form filling, date pickers, and interactive slider adjustments.

% \paragraph{Reliance on Demonstration Trajectory Coverage.}
% The Interaction Element Graph is constructed from successful demonstration trajectories. While aggregating 854 trajectories across 50 websites yields dense coverage of common interaction pathways, novel pages or unexplored site regions are unrepresented in the initial graph topology. Integrating autonomous online exploration agents to dynamically expand the interaction element graph represents a promising direction for continuous environment grounding.

% \paragraph{Integration with Process-Reward Reinforcement Learning.}
% In this work, we evaluated comparative Web PRMs in test-time candidate re-ranking and search. A natural extension is to utilize the comparative PRM as a dense, step-level reward signal in online agent reinforcement learning (e.g., PPO or GRPO) to train policy models directly, guiding exploration without requiring extensive Monte Carlo rollouts.

\end{document}